\documentclass[onefignum,onetabnum]{siamart190516}
\usepackage{cite}
\usepackage{microtype}
\usepackage{graphicx}
\usepackage{subfigure}
\usepackage{booktabs} % for professional tables
\usepackage[normalem]{ulem}

\usepackage{times}
\usepackage{epsfig}
\usepackage{graphicx}
\usepackage{amsmath}
\usepackage{amssymb}
\usepackage{diagbox}

\usepackage{makecell}

\usepackage{amssymb}
\usepackage{setspace}
\usepackage{color}
\usepackage{tabularx}
\usepackage{amsmath}
\usepackage{float}
\usepackage[algo2e,boxed,ruled]{algorithm2e}
\usepackage{fancyhdr}
\usepackage{bm}
\usepackage[utf8]{inputenc} % allow utf-8 input
\usepackage[T1]{fontenc}    % use 8-bit T1 fonts
\usepackage{hyperref}       % hyperlinks
\usepackage{url}            % simple URL typesetting
\usepackage{booktabs}       % professional-quality tables
\usepackage{amsfonts}       % blackboard math symbols
\newcommand{\bfx}{\mathbf{x}}

\newcommand{\CO}{\mathcal{O}}
\newcommand{\cin}{c_{\rm in}}
\newcommand{\cout}{c_{\rm out}}

\usepackage{lipsum}
\usepackage{amsfonts}
\usepackage{graphicx}
\usepackage{epstopdf}
\usepackage{algorithmic}
\ifpdf
  \DeclareGraphicsExtensions{.eps,.pdf,.png,.jpg}
\else
  \DeclareGraphicsExtensions{.eps}
\fi

\newsiamremark{remark}{Remark}
\newsiamremark{hypothesis}{Hypothesis}
\crefname{hypothesis}{Hypothesis}{Hypotheses}
\newsiamthm{claim}{Claim}

\headers{ Multigrid-in-Channels Neural Network Architectures}{M. Eliasof, J. Ephrath, L. Ruthotto and E. Treister}

\title{MGIC: Multigrid-in-Channels Neural Network Architectures\thanks{Corresponding author: Moshe Eliasof.
\funding{The research reported in this paper was supported by the Israel Innovation Authority through Avatar
consortium, and by grant no. 2018209 from the United States - Israel
Binational Science Foundation (BSF), Jerusalem, Israel.  ME
is supported by Kreitman High-tech scholarship. LR's work is partially supported by NSF DMS 1751636.}}}

\author{Moshe Eliasof\thanks{Computer-Science Department, Ben-Gurion University of the Negev.
  (\email{eliasof@post.bgu.ac.il, erant@cs.bgu.ac.il}).}
\and Jonathan Ephrath\footnotemark[2]
\and Lars Ruthotto\thanks{Departments of Mathematics and Computer Science, Emory University.} 
\and Eran Treister\footnotemark[2]
}

\usepackage{amsopn}

\ifpdf
\hypersetup{
  pdftitle={MGIC: Multigrid-in-Channels Neural Network Architectures},
  pdfauthor={M. Eliasof, J. Ephrath, L. Ruthotto ,and E. Treister}
}
\fi

\begin{document}

\maketitle

% REQUIRED
\begin{abstract}
  We present a multigrid-in-channels (MGIC) approach that tackles the quadratic growth of the number of parameters with respect to the number of channels in standard convolutional neural networks (CNNs). Thereby our approach addresses the redundancy in CNNs that is also exposed by the recent success of lightweight CNNs. Lightweight CNNs can achieve comparable accuracy to standard CNNs with fewer parameters; however, the number of weights still scales quadratically with the CNN's width. Our MGIC architectures replace each CNN block with an MGIC counterpart that utilizes a hierarchy of nested grouped convolutions of small group size to address this. 
 Hence, our proposed architectures scale linearly with respect to the network's width while retaining full coupling of the channels as in standard CNNs.
 Our extensive experiments on image classification, segmentation, and point cloud classification show that applying this strategy to different architectures like ResNet and MobileNetV3 reduces the number of parameters while obtaining similar or better accuracy.
\end{abstract}

% REQUIRED

% REQUIRED
%\begin{AMS}
%  68T07, 65N55, 68T45
%\end{AMS}

\section{Introduction}
\label{sec:intro}
Convolutional neural networks (CNNs) \cite{LeCun1990} have achieved impressive accuracy for image classification, semantic segmentation, solution of partial differential equations, and other tasks \cite{krizhevsky2012imagenet,girshick2014rich, khoo2018switchnet}.
The main idea behind CNNs is to define the linear operators in the neural network as convolutions with local kernels.
This increases the network's computational efficiency (compared to the original class of networks) due to the essentially sparse convolution operators and the considerable reduction in the number of weights. 
 The general trend in the development of CNNs has been to make deeper, wider, and more complicated networks to achieve higher accuracy \cite{szegedy2015going}.

In practical applications of CNNs, a network's feature maps are divided into channels, and the number of channels, $c$, can be defined as the width of the layer.
A standard CNN layer connects any input channel with any output channel.
Hence, the number of convolution kernels per layer is equal to the product of the number of input channels and output channels.
Assuming the number of output channels is proportional to the number of input channels, this $\CO(c^2)$ growth of operations and parameters causes immense computational challenges. When the number of channels is large, convolutions are the most computationally expensive part of the training and inference of CNNs. Wide architectures exacerbate this trend with hundreds or thousands of channels, which are particularly effective in classification tasks involving a large number of classes. Increasing the network's width is advantageous in terms of accuracy and hardware efficiency compared to deeper, narrower networks \cite{Zagoruyko:2016wo}.  However, the quadratic scaling causes the number of weights to reach hundreds of millions and beyond \cite{huang2019gpipe}, and the computational resources (power and memory) needed for training and making predictions with such CNNs surpasses the resources of common systems \cite{bianchini2014complexity}.
This motivates the DL community to design more efficient network architectures with competitive performance.

Among the first approaches to reduce the number of parameters in CNNs are the methods of pruning \cite{pruning92,SongHan2015,PrunningCornel2017} and sparsity-promoting \cite{SparsConvCornell2017,SparseReguSongHan2016}, that aimed to limit the connectivity between channels, and have been typically applied to already trained networks. Once a network is trained, a substantial number of its weights can be removed with hardly any degradation of its accuracy. However, the resulting connectivity of these processes are typically  unstructured, which may lead to inefficient deployment of the networks on hardware. Still, pruning serves as a proof-of-concept that the full connectivity between channels is superfluous, i.e., there is a redundancy in CNNs \cite{molchanov2016pruning}. 
By contrast, we reduce the network architecture by using structured convolution operators, to enable efficient, balanced computations during training and inference. 

\begin{figure}
    \centering
    \includegraphics[width=0.9\textwidth]{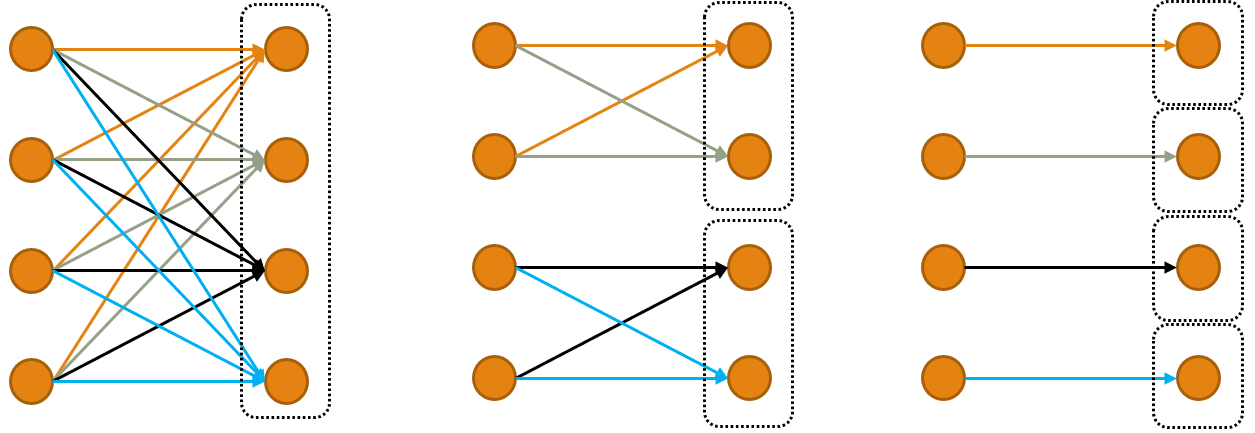}
    \caption{Standard fully-coupled (left), grouped (block-diagonal) convolution operator with 2 groups (middle) and a depth-wise (diagonal) convolution (right).}
    \label{fig:group_conv}
\end{figure}

Another recent effort to reduce the number of parameters of CNNs feature light-weight architectures based on spatial grouped convolutions. The idea is to apply the more computationally expensive spatial convolutions---those involving larger-than-a scalar filters, say, $3\times3$ or $5\times 5$---in small disjoint groups, creating a block-diagonal ``local'' connectivity between channels (see Fig. \eqref{fig:group_conv}). 
The extremity of grouped convolutions (diagonal connectivity) are the so-called ``depth-wise'' convolutions which filter each input channel separately. Such depth-wise convolutions have been commonly used together with point-wise $1\times1$ convolutions, which couple all the channels but with a single scalar for each pair of channel interactions instead of larger stencils like $3 \times 3$ (a $1\times1$ convolution is essentially a simple dense matrix-matrix multiplication). This way, the popular MobileNets \cite{howard2017mobilenets,sandler2018mobilenetv2,howard2019searching} involve significantly fewer parameters than standard networks, while achieving comparable performance. The majority of the weights in MobileNets are in the point-wise operators, which scale with $\CO(c^2)$. The strength of MobileNets, and its improvement EfficientNet \cite{tan2019efficientnet}, is the inverse bottleneck structure, that takes a narrow network (with relatively few channels) and expands it by a significant factor to perform the depth-wise and non-linear activation. This way, although the number of parameters scales quadratically in the width in the $1\times 1$ operators, it aims to maximize the spatial convolutions and activations as much as possible to increase the expressiveness of the network. The ShuffleNet~\cite{ma2018shufflenet} reduces the parameters of the point-wise operator by applying $1\times1$ convolutions to half of the channels and then shuffling them. Another closely related architecture include LeanConvNets \cite{ephrath2020leanconvnets} which employs 3 and 5 points convolution stencils to reduce computational complexity, together with grouped convolutions. The recent GhostNet \cite{han2020ghostnet} divides the feature space by some hyper parameter (typically, 2 or 4), such that only a subset of the channels are convolved by a $1\times 1$ convolution, and the remaining channels are obtained from the densely-learned portion of the feature space by depth-wise convolutions.

\begin{figure*}
    \centering
    \includegraphics[width=1.0\textwidth]{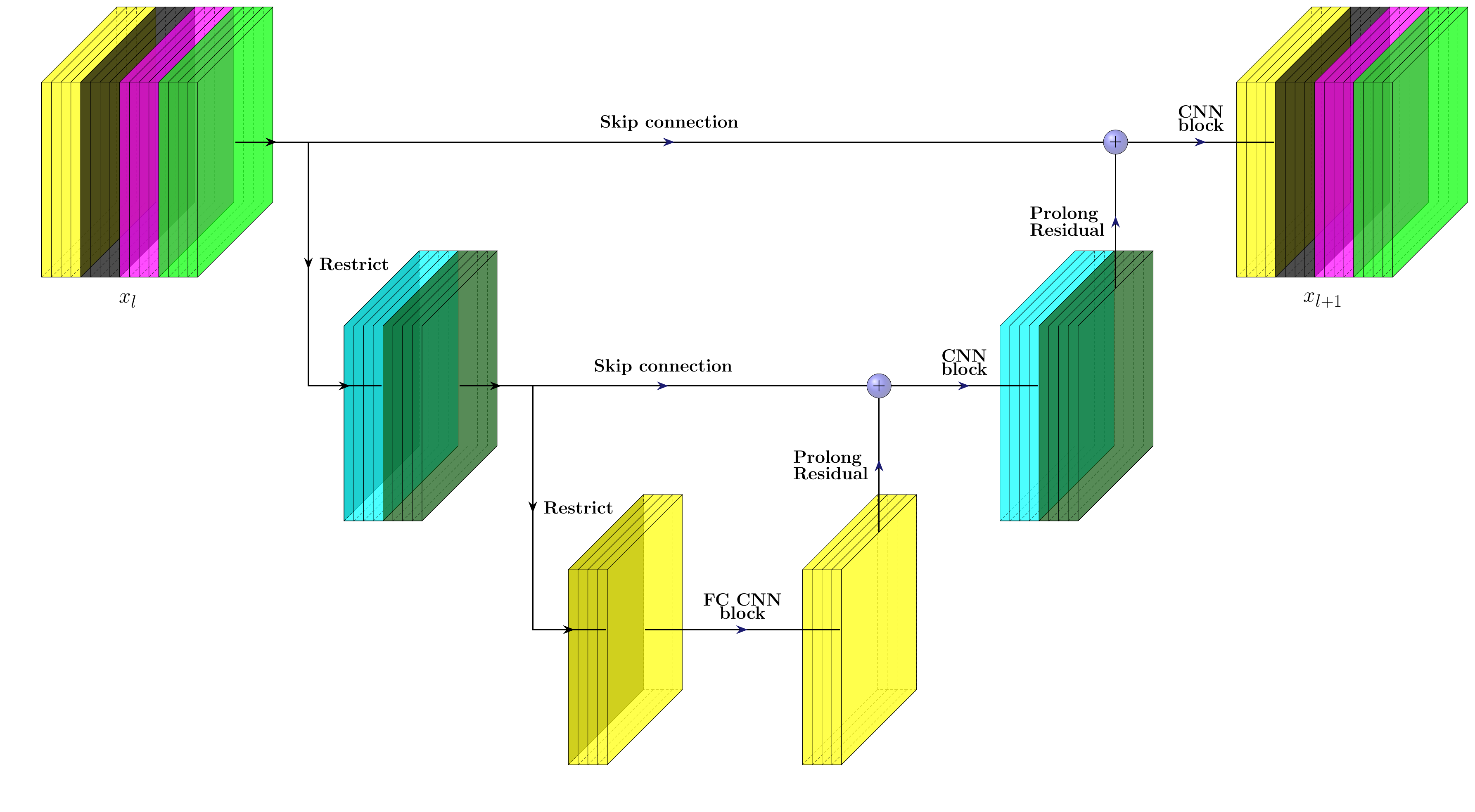}
\caption{A three-level multigrid block for 16 input channels and a group size $s_g=4$. \emph{Restrict} and \emph{Prolong Residual} denote grid transfer operators, which decrease and increase the number of channels, respectively. All the channels in the block are of the same spatial resolution. Each color denotes a group of channels that are mixed in a CNN block. The coarsest level uses a fully coupled CNN block. 
    }
    \label{fig:MG-in-channels}
\end{figure*}

In this work, we aim to obtain full connectivity between the channels of convolutional neural networks using locally grouped convolutions \textit{only}. To obtain this, we adopt the multigrid approach \cite{briggs2000multigrid} that is known to be effective in solving various large problems using local processing only on a hierarchy of grids or meshes. Such multigrid methods are primarily used to solve differential equations and other graphical problems (e.g., Markov chains, \cite{de2008multilevel}). They are based on a principle that a local iterative solution process on a fine grid can effectively eliminate errors that involve short distances on the grid (also known as, ``smoothing'') but cannot reduce long-distance errors. Such errors can be approximated on a coarser grid, leading to two advantages. First, coarse grid procedures are less expensive than fine grid procedures as they involve fewer grid points.
Second, traversing different scales leads to faster convergence. An alternative interpretation of this paradigm is that the multigrid hierarchy efficiently transfers information across all the grid points using local processing only, at a different scale at each level. Classical multigrid methods rely on the multiscale representation of functions in space but can also be used to tackle temporal problems \cite{falgout2014parallel}.  
Multigrid has been abundantly
applied in the context of partial differential equations \cite{briggs2000multigrid}, but also in
other areas such as optimization \cite{brandt2003multigrid,borzi2009multigrid,treister2016multilevel}, Markov chains \cite{de2008multilevel}, and
graphs \cite{livne2012lean,napov2017efficient}. Multigrid approaches have also been used to warm-start the training of CNNs on high-resolution images with training on low-resolution images \cite{haber2018learning}, adopting a multiscale approach in space.
Similarly, \cite{MultigridNN, pelt2018mixed, he2019mgnet} define multiscale architectures that extract and combine features from different resolution images scales.
The DeepLabV3 architecture for semantic segmentation \cite{chen2017rethinking} also exploits multiscale representations. Other works \cite{chen2018big, Chen:2019tp, wang2019elastic} present different strategies to exploit the spatial multiscale structure and representation of the feature maps throughout CNNs to improve the performance of standard networks.
Multigrid has also been used in the layer dimension for residual networks, e.g., to warm-start the training of a deep network by interpolating the weights of a trained shallow network with a larger step size  \cite{chang2017multi}, and improving or parallelizing the training through the layers~\cite{guenther2020layer,kopanivcakova2021globally}. While the mentioned works apply the multigrid idea either in space or in layers (depth), in this work we use the multigrid idea in the channel space (width).

In this work we propose a \textit{multigrid-in-channels} (MGIC) approach for building network architectures that connect all channels at each layer
using local grouped convolutions only. We achieve this full coupling via a hierarchical structure of the features in the channel-space, and grouped convolutions with relatively small groups of fixed size at each level of the hierarchy. This yields a computational complexity that scales linearly with the network's width. That is, for a fixed group size, doubling the number of channels in the network doubles the number of parameters. This allows us to use wider, deeper, and ultimately more expressive networks for a given computational budget.

Our MGIC block, which we show in  Fig. \ref{fig:MG-in-channels}, can replace standard convolutional blocks in CNNs. At each level of the block, a particular width is considered in the channel space. Coarser levels in the block are defined by clusters of channels formed by averaging channels from different groups and therefore have fewer channels. On each level, we apply grouped convolutions on those clustered channels, which effectively connect different fine-level groups.
We also note that other forms of local convolutions, like a chain or two-dimensional lattice structure, are also suitable in our framework, as long as they are of linear complexity. We chose grouped convolutions as they are the only type of local convolution in the channel space that is supported in deep learning frameworks.

Even though our MGIC block relies on local grouped convolutions, its multigrid structure enables communication between all channels. Our experiments on feature--space representation using a MGIC block, as well as implicit function approximation suggest that our method is of high capacity and expressiveness achieved in a linear scaling fashion. Furthermore, our experiments on image classification, segmentation, and point cloud classification tasks, the MGIC-block achieves competitive or superior performance with a relatively low number of parameters and FLOPs.

%------------------------------------------------------------------------
\section{Towards efficient convolutions}
Typical CNN architectures are composed of a series of blocks
\begin{equation}\label{eq:CNN-block}
\mathbf{x}_{l+  1} =  \mbox{CNN-block}(\mathbf{x}_l), 
\end{equation}
where $\mathbf{x}_l$ and $\mathbf{x}_{l+1}$ are the input and output features of the $l$-th block, respectively. Each CNN-block usually contains a sequence of basic layers, with associated weights which are omitted in the following. For example, consider the bottleneck ResNet block \cite{he2016deep} that reads
\begin{equation}\label{eq:bottleneck}
\mathbf{x}_{l+1} =\mathbf{x}_l +K_{l_3}\sigma(\mathcal{N}(K_{l_2}(\sigma(\mathcal{N}(K_{l_1}\sigma(\mathcal{N}(\mathbf{x}_l)))).
\end{equation}
The operators $K_{l_1}$ and $K_{l_3}$ are fully coupled $1\times1$ convolutions, and $K_{l_2}$ is a $3\times3$ convolution. 
The function $\sigma()$ is a non-linear point-wise activation function, typically the ReLU function $\sigma (x)=\max (x,0)$. $\mathcal{N}$ is a normalization operator often chosen to be a batch normalization. Architectures that follow a similar trend are MobileNets \cite{howard2017mobilenets,sandler2018mobilenetv2,howard2019searching} and ResNeXt \cite{xie2017aggregated}, in which 
$K_{l_2}$ is a grouped (or depth-wise) convolution to reduce the number of parameters and increase the ratio between activations and parameters. 

A convolution layer takes $c_{\rm in}$ channels of feature maps, and outputs $c_{\rm out}$ such channels. By definition, a fully coupled convolution is involved with $\CO({c_{\rm in}\cdot c_{\rm out}})$ parameters and FLOPs, i.e., both scale quadratically with the width. Since popular implementations of CNNs often use hundreds or thousands of channels, their full coupling leads to large computational costs and to millions of parameters for each layer, which may not be necessary.
 
To ease the computational costs, all or some of the convolution layers in the CNN block \eqref{eq:CNN-block} may be grouped, dividing the input channels to several equally-sized groups and applying a separate convolution kernel on each of them. In this work, we denote the number of channels in a group (its size) by $s_g$. For example, a standard fully-coupled convolution with one group is defined by $s_g=c_{\rm in}$ where $c_{\rm in}$ is the number of input channels, while a depth-wise convolution is achieved by $s_g=1$. As the group size gets smaller, the grouped convolutions involve less computations, at the cost of typically less expressive network architectures. That is because information cannot be shared between different groups in the feature maps during the grouped convolutions.  

In this work, we propose to replace the $\mbox{CNN-block}$ in~\eqref{eq:CNN-block}  by a novel multigrid block to obtain the forward propagation
\begin{equation}\label{eq:MG-block}
\mathbf{x}_{l+1} = \mbox{MGIC-block}(\mathbf{x}_l,\mbox{CNN-block},s_g,s_c),
\end{equation}
which, as illustrated in Fig. \ref{fig:MG-in-channels}, uses a hierarchy of grids in the channel space and applies the original CNN-block on the coarsest level. 
The parameter $s_g$ defines the group size of the convolution operators in these CNN-blocks, and $s_c$ is the size of the coarsest grid. As we show in Sec. \ref{sec:complexity}, the number of parameters and FLOPs in the MGIC-block scale \textit{linearly} with respect to the number of channels, assuming that the group size is fixed. Note that the MGIC block is agnostic to its CNN-block, and therefore can be used for various CNN architectures, including future ones.

%------------------------------------------------------------------------

\section{Multigrid-in-channels CNN architectures}
\label{sec:proposedMethod}
In this section, we describe the MGIC-block in detail. We start by defining the multigrid hierarchy. Then, we define the MGIC-block and the grid transfer operators, which are essential to perform down and up sampling of the channel space, followed by a comprehensive description of the design of such MGIC-block. Finally, we compare the computational cost of a standard CNN layer with ours.

\subsection{The multigrid hierarchy}
The key idea of our multigrid architecture is to design a hierarchy of grids in the channel space (also referred to as ``levels''), where the number of channels in the finest level corresponds to the original width of the network. The number of channels is halved between the levels until reaching the coarsest level, where the number of channels is smaller or equal to the parameter $s_c$. Our multigrid architecture is accompanied by a CNN block, like the ResNet block in Eq. \eqref{eq:bottleneck}, which is applied on each level. On the finest and intermediate levels, we only connect disjoint groups of channels using grouped convolutions. These convolutions have $\CO(s_g \cdot c_{\rm in})$ parameters, and we keep the group size $s_g$ fixed throughout the network. Hence, as the network widens, the number of groups grows, and the number of parameters grows linearly. We allow interactions between all the channels on the coarsest grid, where we use the original CNN-block without grouping. Since the coarsest grid contains only a few channels, this is not a costly operation. 
We note, that our architecture is capable of performing more convolution layers and non-linear activations per MGIC-block, which is designed to replace a given CNN-block. That is, our MGIC approach can yield models with higher capacity and expressiveness while retaining similar or lower computational cost, due to the use of grouped variants of the original, fully-coupled CNN block.

\subsection{The multigrid block}\label{sec:MGBlock}
For simplicity, we assume that the CNN-block and the MGIC-block change neither the number of channels nor the spatial resolution of the images. That is, both $\bfx_l$ and $\bfx_{l+1}$ in \eqref{eq:MG-block} have $c_{\rm in}$ channels of the same spatial resolution. Given a CNN-block, a group size $s_g$ and a coarsest grid size $s_c$, we define the multigrid block in Alg. \ref{alg:MGStep}, and as an example we present and discuss a two-level version of it below. Here, the two-level hierarchy is denoted by levels $0,1$, and $\bfx^{(0)} = \bfx_l$ are the input feature maps at the finest level (level $0$). The two-level block is as follows
\begin{eqnarray}
\bfx^{(1)}  &=& R_0\bfx^{(0)} \label{eq:MulRestriction}\\
\bfx^{(1)} &\leftarrow& \mbox{CNN-block}(\bfx^{(1)}) \label{eq:Coarsest}\\
 \bfx^{(0)} &\leftarrow& \bfx^{(0)}+\mathcal{N}(P_0(\bfx^{(1)}-R_0\bfx^{(0)}))
 \label{eq:InterpolatedGrid}\\
\mathbf{x}_{l+1} &=& \mbox{CNN-block}(\bfx^{(0)},s_g) \label{eq:fineCNN}
\end{eqnarray}
We first down-sample the channel dimension of the input feature maps $\bfx^{(0)}$ in Eq. \eqref{eq:MulRestriction} by a factor of 2, using a restriction operator $R_0$. This operation creates the coarse feature maps $\bfx^{(1)}$, which have the same spatial resolution as $\bfx^{(0)}$, but half the channels. The operator $R_0$ is implemented by a grouped $1\times1$ convolution; see a detailed discussion in Sec. ~\ref{sec:transfer}. Then, in Eq. \eqref{eq:Coarsest} a non-grouped CNN block is applied on the coarse feature maps $\bfx^{(1)}$. This block couples all channels but involves only $c_{\rm in}^2/4$ parameters instead of $c_{\rm in}^2$. Following that, in Eq. \eqref{eq:InterpolatedGrid} we use a prolongation operator $P_0$
to up-sample the residual $\bfx^{(1)} - R_0\bfx^{(0)}$\footnote{In the multigrid literature, this term is called a coarse grid correction, because it corrects the the fine level solution using an interpolated coarse error approximation. Here, there is no iterative solution, as we just propagate feature maps through the network's layer. Hence, we use the term residual that is common in deep learning literature as the addition to the feature maps in a layer.} from the coarse level to the fine level (up-sampling in channel space) and obtain a tensor with $c_{\rm in}$ channels. Adding the up-sampled residual is common in non-linear multigrid schemes---we elaborate on this point below.
Finally, in Eq. \eqref{eq:fineCNN} we perform a grouped CNN block, which is of significantly lower computational cost than its non-grouped counterpart for a small $s_g$. An illustration of this architecture using three levels is presented in Fig. \ref{fig:MG-in-channels}. The multilevel block (Alg. \ref{alg:MGStep}) is applied by iteratively reducing the channel dimensionality until reaching the coarsest grid size $s_c$. Hence, the number of levels at each layer is given by $\textstyle{n_{\rm levels} = \Bigl\lfloor\log_2(\frac{c_{\rm in}}{s_c})\Bigr\rfloor}$, and the architecture uses more levels as the channel space widens. By choosing $s_g = s_c$ we have grouped convolutions on the fine and intermediate levels and a fully connected layer only on the coarsest level, which is a natural configuration that is also illustrated in Fig. \ref{fig:MG-in-channels}. Specifically, in our experiments we choose $s_c = s_g$ or $s_c = 2s_g$ which can serve as default choices.

By letting the information propagate through the multigrid levels sequentially, we increase the number of convolutions and non-linear activations compared to the original CNN block. Thereby, we aim to increase the expressiveness of our MGIC-block, given a computational budget. 

\begin{algorithm2e}
\DontPrintSemicolon
\KwSty{Algorithm:} \\ $\bfx_{l+1}= \mbox{\textbf{MGIC-block}}(\bfx_l,\mbox{CNN-block},\mathit{s_{g}},\mathit{s_c})$.\;
\emph{\# Inputs: \\
\# $\bfx_l$ - input feature maps with $c_{\rm in}$ channels. \\ \# $s_{g}$: group size. $s_c$: coarsest grid size. \\ \# CNN-block: A reference CNN block, e.g., Eq. \eqref{eq:bottleneck}}\;
$\bfx^{(0)} = \bfx_l$  \;
$n_{\rm levels} = \Bigl\lfloor\log_2(\frac{c_{\rm in}}{s_c})\Bigr\rfloor$\;
\emph{\# Going down the levels, starting from $\bfx_l$}\;
\For{$j=0:n_{\rm levels}$}{
$\bfx^{(j+1)} = R_j\bfx^{(j)}$\;
}
\emph{\# On the coarsest level we perform a non-grouped  CNN-block:}\;
$\bfx^{(n_{\rm levels})} \leftarrow  \mbox{CNN-block}(\bfx^{(n_{\rm levels})})$\;
\emph{\# Going up the levels:}\;
\For{ $j=n_{\rm levels}-1:0$}{
$\bfx^{(j)} \leftarrow \bfx^{(j)} + \mathcal{N}\left(P_j(\bfx^{(j+1)} - R_j\bfx^{(j)})\right)$\;
$\bfx^{(j)} \leftarrow \mbox{CNN-block}(\bfx^{(j)}, \mbox{group\_size} = s_{g})$\;
}
return $\bfx_{l+1} = \bfx^{(0)}$.
\caption{Multigrid-in-channels block }\label{alg:MGStep}
\end{algorithm2e}

\subsection{The choice of transfer operators $P$ and $R$}\label{sec:transfer}
The transfer operators play an important role in multigrid methods. In classical methods, the restriction $R$ maps the fine-level state of the iterative solution onto the coarse grid, and the prolongation $P$ acts in the opposite direction, interpolating the coarse solution back to the fine grid. Clearly, in the coarsening process we lose information, since we reduce the dimension of the problem and the state of the iterate. The key idea is to design $P$ and $R$ such that the coarse problem captures the subspace that is causing the fine-grid process to be inefficient. This results in two complementary processes: the fine-level steps (dubbed as \emph{relaxations} in multigrid literature), and the coarse grid correction.

To keep the computations low, at the $j$-th level we choose $R_j$ to be a grouped $1\times1$ convolution that halves the number of channels of its operand. We choose $P_j$ to have the transposed structure of $R_j$. For $R_j$ and $P_j$ we choose the same number of groups as in the CNN-block 
, e.g., for $R_0$ it will be $\frac{c_{\rm in}}{s_g}$ groups. The operators $R_j$ take groups of channels of size $s_g$ and using a $1\times1$ convolution distills the information to $\frac{s_g}{2}$ channels. Similarly, the operators $P_j$ interpolate the coarse channels back to the higher channel dimension using grouped $1\times1$ convolutions. This choice for the transfer operators corresponds to aggregation-based multigrid coarsening \cite{vanvek1996algebraic}, where aggregates (groups) of variables are averaged to form a coarse grid variable. In Sec.  \ref{sec:representation_experiment},  we exemplify that the transfer operators preserve essential information of all channels.

The weights of the transfer operators are learned as part of the optimization, and are initialized by positive weights with row-sums of 1. This initialization ensures that feature maps do not vanish as we multiply them by consecutive restrictions $R_j$ to start the MGIC-block (in the first loop of Alg. \ref{alg:MGStep}) at the beginning of the training process. From this initialization we optimize the weights during training. This is also motivated by similar works on trained optimization algorithms~\cite{HochreiterLearning} and their success in imaging applications~\cite{ hammernik2018learning}. 

\subsubsection*{The importance of up-sampled residuals}
Adding the up-sampled residual in \eqref{eq:InterpolatedGrid} is often called a $\tau$-correction in multigrid literature, and it is the standard way to apply multigrid methods to solve non-linear problems \cite{briggs2000multigrid}. Here, it allows us to have a skip connection between corresponding levels of the multigrid cycle, introducing an identity mapping in \eqref{eq:InterpolatedGrid} as guided by \cite{he2016deep}. Furthermore, at each level $j$ we up-sample only the difference between feature maps  $\bfx^{(j+1)}-R_j\bfx^{(j)}$ and not the feature map $\bfx^{(j+1)}$ itself. This helps preventing feature maps summation when adding the up-sampled residual, which may lead to exploding gradients.
By this definition, if the CNN-block has an identity mapping, then so does the whole MGIC-block in Alg. \ref{alg:MGStep}.

\subsubsection*{Changing the channel resolution blocks}
The structure of the MGIC block as in Fig. \ref{fig:MG-in-channels} is more natural to equal input and output channel sizes, i.e., $c_{\rm in} = c_{\rm out}$. Hence, when we wish to change the number of channels , we define a lightweight shortcut that is designed to transform a tensor from $\cin$ to $\cout$ such that our MGIC-blocks will be given an input where $\cin = \cout$. Specifically, to obtain low computational cost, we use a depth-wise $3 \times 3$ convolution, although other alternatives such as a $1\times1$ convolution are also possible. In case we wish to change the spatial dimensions of the input tensor, we perform the same operation, only with a stride of 2.

\subsection{The complexity of the MGIC-block}\label{sec:complexity}
Consider a case where we have $c = c_{\rm in} = c_{\rm out}$ channels in the network, and we apply a standard convolution layer using $d\times d$ convolution kernels  (e.g., a $3\times 3$ kernel). The output consists of $c$ feature maps, where each one is a sum of the $c$ input maps, each convolved with a kernel. Hence, such a convolution layer requires $\CO(c^2 \cdot d^2)$ parameters, inducing a quadratic growth in the parameters and FLOPs.

\subsubsection*{Relaxation cost per level} On each level of an MGIC block, a relaxation step is performed. At the $j$-th level, this relaxation step is realized by a grouped convolution of kernel size $d \times d$, with a group size of $s_g$ that divides  $\frac{c}{2^j}$ (since at each level we halve the channels space, starting from $c$ channels), yielding $\frac{c}{s_g\cdot 2^j}$ groups. Therefore, the number of parameters required for such relaxation step is $ \frac{s_g\cdot c \cdot d^2}{2^j}$. At the coarsest level, we have $s_c$ channels which perform a fully-coupled relaxation step, requiring $s_c \cdot d^2$ parameters.

\subsubsection*{The cost of restriction and prolongation}
As discussed in Sec. \ref{sec:transfer}, the restriction and prolongation operators are implemented via grouped $1\times1$ convolutions, halvening and doubling the feature space dimension, respectively. Those operators are learned  at each level of our MGIC block. Therefore, the number of parameters for those operators at the $j$-th level is  $\frac{c}{2^j}$. The analysis here is similar to the case of the relaxtion steps, only here $d=1$, and we have no fully-coupled operators on the coarsest level.

\subsubsection*{The total cost of an MGIC block}
Combining the analysis from the paragraphs above, the total number of parameters for an MGIC block with $n$ levels is as follows:
\begin{equation}\ \sum_{j=0}^{n-1} \left(\frac{s_g \cdot c\cdot (d^2+1)}{2^j}\right) + s_c^2\cdot d^2 < 2\left(s_g \cdot c\cdot (d^2+1)\right) + s_c^2\cdot d^2.
\end{equation}
If $s_c$ is small (typically, we choose $s_c = s_g$) , we can neglect the term $s_c^2\cdot d^2$ to obtain $\CO(s_g \cdot c\cdot (d^2+1))$ parameters. Therefore, since $s_g$ and $s_c$ are fixed and small, and the spatial dimension of the learned relaxation step convolution kernel size $d$ is typically of small size (3, 5, or 7), our method scales linearly with respect to the network's width. This will be most beneficial if $c$ is large, which is typical and usually required in order to obtain state-of-the-art performance on various tasks \cite{he2016deep,brown2020languagegpt}  as in discussed in Sec. \ref{sec:intro}.

\subsubsection*{Memory footprint} During training, the memory footprint of MGIC is roughly twice as large as that of a single CNN block since all the maps in the hierarchy are saved for backpropagation. However, the main motivation for using light-weight networks is increasing the efficiency of the inference application. This is beneficial when the trained models are  applied over and over again for inference, possibly on edge devices with fewer computational resources, like autonomous cars and drones.
This typically offsets the increased cost of training, which is typically done once or a few times.  Indeed, during inference the coarser feature maps are released while going up the hierarchy, so when applying the upmost CNN block, the memory footprint is identical to a single block. Following the complexity analysis above, all the feature maps $\{\textbf{x}^{(l)}\}$ require about $\times 2$ the memory of $\textbf{x}^{(0)}$, but the memory footprint of a CNN block can be higher than that. For example, the MobileNets\cite{howard2019searching} involve an inverse bottleneck with an expansion of $4-6$,  rendering it more expensive in terms of peak-memory as the additional MGIC overhead.

%---------------------------------------------------------------------

\section{Experiments}
\label{sec:experiments}
In this section, we report several experiments with our MGIC approach. We start with two proof-of-concept experiments, measuring how good our MGIC can compress the channel space, as well as its capacility of approximating implicit functions. Then, we test our method on image classification and segmentation and point cloud classification benchmarks. Our goal is to compare how different architectures perform using a relatively small number of parameters, aiming to achieve similar or better results with fewer parameters and FLOPs.
We train all our models using an NVIDIA Titan RTX and implement our code using the PyTorch software \cite{paszke2017automatic}. Our code is available at: \url{github.com/BGUCompSci/MultigridInChannelsCNNs}
%----------------------------------------------------------------------

\subsection{Coarse channels representation}\label{sec:representation_experiment}
\begin{figure}
    \centering
    \includegraphics[width=0.9\textwidth]{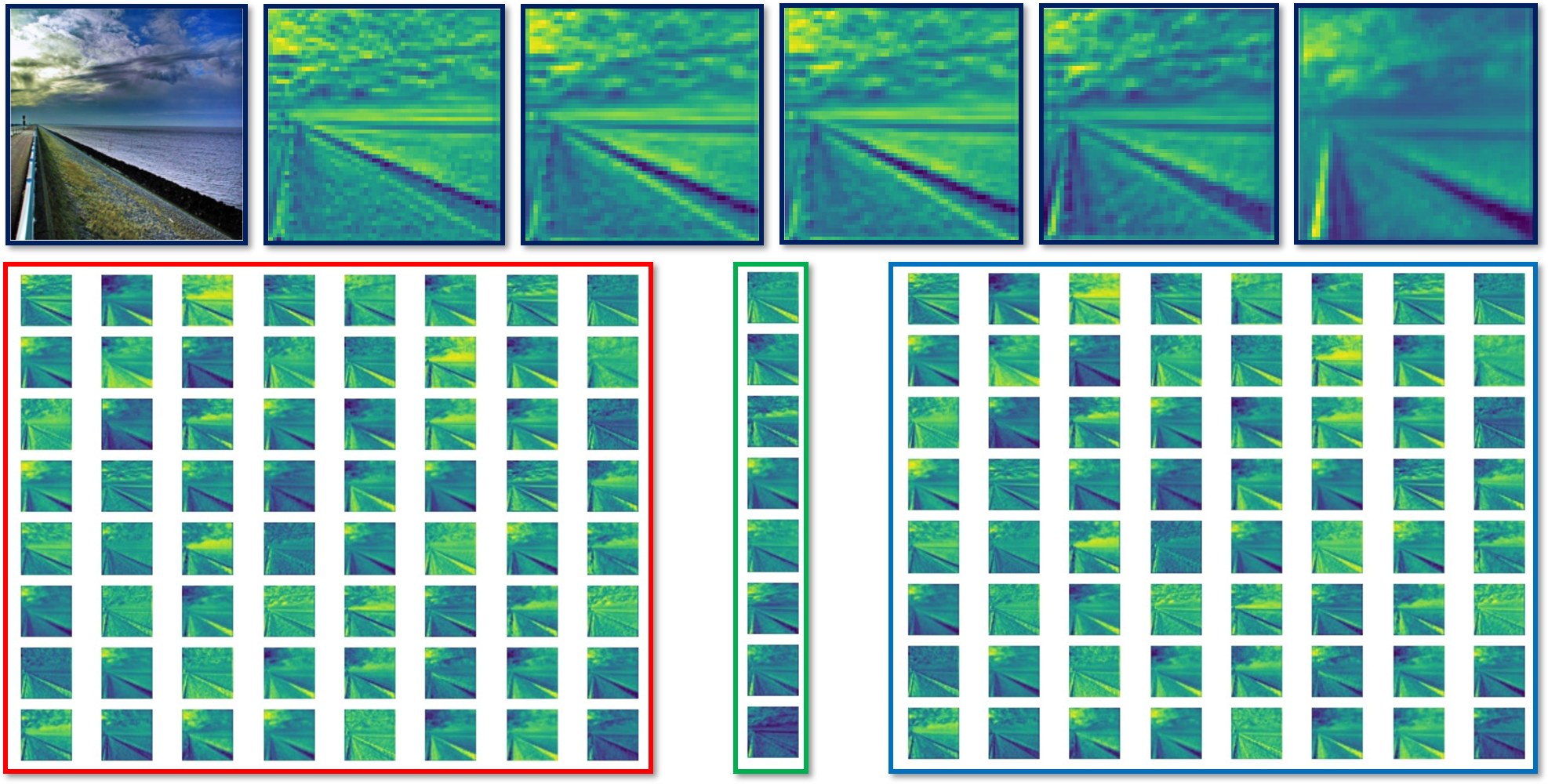}
    \caption{From left to right:  Top - input image, one of its feature maps, and its reconstructions with $s_g=32, 16, 8, 4$, respectively.  $s_c$ is kept at $8$. Bottom - input feature maps, their coarsest grid representation, and their reconstruction from our MGIC block with down and up-sampling only.}
    \label{fig:reconstruction}
\end{figure}

Our method aims to reduce parameters by introducing a hierarchical representation of the network's channels, and traverse this hierarchy in the forward pass. In this experiment we wish to quantify the effectiveness of our channel down and up sampling mechanism. Specifically, we wish to measure how well we can encode feature maps on the coarsest grid, where fully coupled convolutions are applied, using the transfer operators $R$ and $P$ only. To do so, we sample $1,024$ images from ImageNet, and extract their feature maps from the first convolutional layer of a pre-trained ResNet-50, containing 64 channels. Then, we encode and decode the feature maps using the restriction and prolongation operators, respectively. To study the transfer operators in isolation, we remove the CNN blocks and long skip connections in Fig. \ref{fig:MG-in-channels} from the MGIC-block. This experiment disentangles the concept of coarse channel representation used in MGIC from the actual CNN-block that is used in the other experiments that we show in this paper. We experiment with several values of the group size parameter $s_g$ and present the mean squared error of the feature maps reconstruction in Tab. \ref{tab:mse}. The original feature maps and their reconstructions in Fig. \ref{fig:reconstruction}. According to this experiment, our method is capable of faithfully representing the original channel space (obtaining low MSE values), even when it operates on a low parameters and FLOPs budget (the values of $s_g$ and $s_c$ are small compared to 64 channels). Obviously, as we use a larger group size, the approximation improves, but not dramatically---it is only a factor of about 2x in the MSE between $s_g=32$ and $s_g=4$.

\begin{table}
  \caption{Feature maps reconstruction mean square error (MSE) v.s. $s_g$. $s_c$ is fixed to $8$. The numbers below are rounded.}
  \label{tab:mse}
  \centering
 \begin{tabular}{lcccc}
    \toprule
    $s_g$  & 32 & 16  &  8 & 4 \\
    \midrule
    \midrule
    MSE &  0.011 & 0.013 & 0.017 & 0.024 \\
    Parameters & 3,300 & 1,800 & 900 & 450 \\
    \bottomrule
  \end{tabular}
\end{table}

%-----------------------------------------------------------------------

\subsection{Performance in function approximation}\label{sec:implicit_reconstruction_exp}
\begin{figure}
    \centering
    \includegraphics[width=0.55\textwidth]{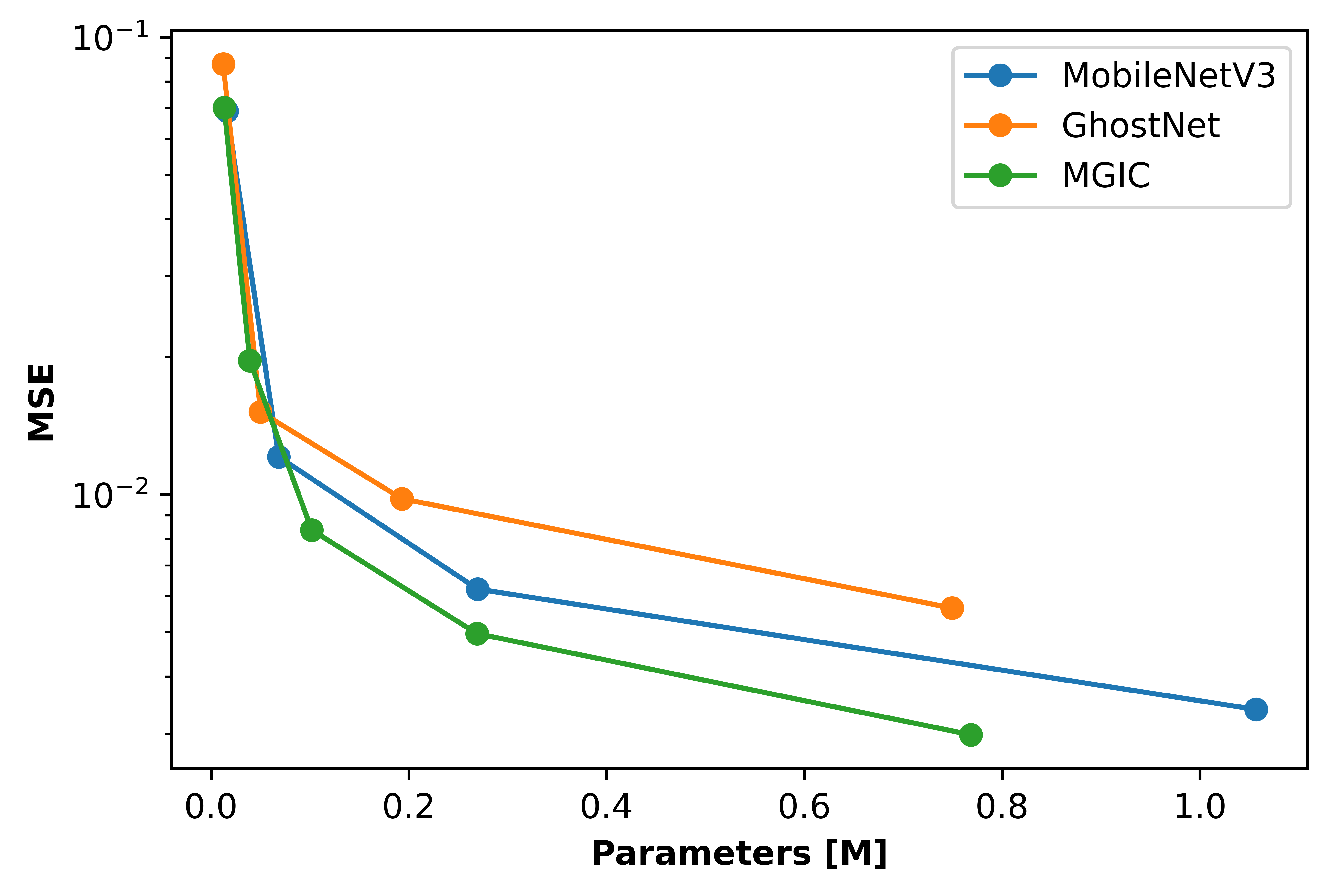}
    \caption{Approximation error of the function family from Eq. \eqref{eq:approxfuncfamily} with MobileNetV3, GhostNet and our MGIC. Metric is in MSE as function of number of parameters.
    }
    \label{fig:implicitRecons}
\end{figure}

In this experiment we demonstrate the efficacy in function approximation in a supervised learning setup. This property is important, especially in applications where we wish to model implicit functions via a neural networks, such as signed-distance fields for shape reconstruction and completion \cite{park2019deepsdf}, and solution of PDEs \cite{pang2019fpinns, raissi2019physics,bar2019unsupervised,lu2020deepxde}---these works in particular approximate functions in an unsupervised manner. Here, we wish to approximate a family of functions

\begin{equation}
\label{eq:approxfuncfamily}
f(x,y,a,b,c) = a\cos(bx) \sin(cy),
\end{equation}
where $x,y\in[0,2\pi] \ , a \in [0,1] \ , b \in [1,2] \ , c \in [10,20]$. We use three methods -- MobileNetV3 \cite{howard2019searching}, GhostNet \cite{han2020ghostnet} and our MGIC-MobileNetV3. Since the input is a vector $[x,y,a,b,c]\in \mathbb{R}^5$ (treated as a spatial domain of size 1), the convolution kernels are effectively $1\times 1$ convolutions. (Note that in any case these are the dominant operations in all the networks and are the driving force of neural network in general.) Throughout all the experiments, we used the network described in Tab. \ref{tab:archImplicit}, where CNN-block is replaced with the respective method. The architecture takes a hyper-parameter $c_{max}$, which dictates the maximal width of the network. $c_{max}$ is assumed to be a power of $2$ and larger than $32$. Furthermore, we use two CNN-blocks for each width. For instance, when $c_{max} = 128$, there are a total of six CNN-blocks of the following widths $\{32,32,64,64,128,128\}$ (these are the $c_{out}$ values). The settings of this experiments are as follows: we randomly sample $10$ million points of the functions defined by $f$ in \eqref{eq:approxfuncfamily}, where  $95 \%$ of the points are used for training, and the remaining $5 \%$ for testing, with a 10-fold cross-validation. We train each network for $1,000$ epochs with a batch size of $20,000$ points, using the SGD optimizer with a constant learning rate of $0.0001$. The loss function is the mean squared error (MSE).
The results, summarized in Fig. \ref{fig:implicitRecons}, show that MGIC yields lower MSE with fewer parameters than MobileNetV3 and GhostNet, suggesting that at least for the purpose of approximating such functions, our MGIC architecture has a higher model capacity per number of parameters.

\begin{table}
  \caption{Network architecture used in the function approximation experiment \ref{sec:implicit_reconstruction_exp}. -- denotes a non-applicable parameter. BN denotes a batch-normalization operator. For CNN-block, we consider and compare MobileNetV3, GhostNet and our MGIC-MobileNetV3. $c_{max}$ is the maximal number of channels (e.g., $128$)}
\centering
  \label{tab:archImplicit}
 \begin{tabular}{lccc}
    \toprule
    $c_{in}$ & Operations & Expansion  & $c_{out}$ \\
    \midrule
    \midrule
    $5 $ &  $1\times 1$ Conv, BN, ReLU & --  & $16$ \\
    $16 $ &  CNN-block & $4$  &  $32$ \\
    $32$  &   CNN-block & $4$  & $32$ \\
    $32$  &   CNN-block & $4$  & $64$ 
    \\
    &  \vdots &
    \\
    $c_{max}$  &   CNN-block & $4$  & $c_{max}$ \\
    $c_{max}$  &   CNN-block & $4$  & $c_{max}$ \\
    $c_{max}$ &  $1\times 1$ Conv, BN, ReLU  & --  &$64$ \\
    $64 $ &  $1\times 1$ Conv & -- & $1$ \\
    \bottomrule
  \end{tabular}
\end{table}

\iffalse
\begin{table}
  \caption{Network architecture used in the implicit function representation experiment. -- denotes a non-applicable parameter. $\alpha$ denotes a width-multiplier. BN denotes a batch-normalization operator. CNN-block can be any block (e.g., MGIC-block). $n$ denotes the number of points.}
  \centering
  \label{tab:archImplicit}
 \begin{tabular}{lccc}
    \toprule
    Input & Operations & Expansion  & $c_{out}$ \\
    \midrule
    \midrule
    $n \times 2 $ &  $1\times 1$ Conv, BN, ReLU & --  & 16 \\
    $n \times 16 $ &  CNN-block & 2  & $\alpha \cdot  160$ \\
    $n \times  \alpha \cdot  160$  &   CNN-block & 2  & $\alpha \cdot 160$ \\
    $n \times  \alpha \cdot  160 $ &  $1\times 1$ Conv, BN, ReLU  & --  &64 \\
    $n \times 64 $ &  $1\times 1$ Conv, BN, ReLU & -- & 2 \\
    \bottomrule
  \end{tabular}
\end{table}
\fi

%-----------------------------------------------------------------------
\subsection{Image classification}
We compare our approach with a variety of popular and recent networks like ResNet-50 \cite{he2016deep}, MobileNetV3 \cite{howard2019searching} and GhostNet \cite{han2020ghostnet} for image classification on the CIFAR10 and ImageNet datasets. 
 We use SGD optimizer with a mini-batch size of $256$ for ImageNet, and $128$ for CIFAR-10, both for $100$ epochs. Our loss function is cross-entropy. The initial learning rates for CIFAR-10 and ImageNet are $0.001$ and $0.1$, respectively. We divide them by $10$ every $30$ epochs. The weight decay is $0.0001$, and the momentum is $0.9$. As data augmentation, for both datasets, we use standard random horizontal flipping and crops, as in \cite{he2016deep}.

\subsubsection{CIFAR-10}

The CIFAR-10 dataset \cite{krizhevsky2009learning} consists of 60K natural images of size $32\times32$ with labels assigning each image into one of ten categories. The data is split into 50K training and 10K test sets. Here, we use a ResNet-$56$ \cite{he2016deep} architecture, which includes ResNet layers of widths 16, 32 and 64 channels. Together with our MGIC block, with parameters $s_g = 8,\; s_c=16$, MGIC-ResNet-56 has up to three levels.
We compare our method with other recent and popular architectures such as AMC-ResNet-$56$ \cite{he2018amc} and Ghost-ResNet-$56$ \cite{han2020ghostnet}, and our baseline is the original ResNet-$56$. We report our results in Tab. \ref{tab:cifarCompression}, where we see large improvement over existing methods, while retaining low number of parameters and FLOPs.

\begin{table}[ht]
  \caption{Comparison of state-of-the-art methods for compressing
 ResNet-$56$ on CIFAR-$10$. - indicates unavailable results. }
  \label{tab:cifarCompression}
  \centering
  \begin{tabular}{lcccc}
    \toprule
    Architecture & \makecell{Params $[\times 10^6]$} & \makecell{FLOPs $[\times 10^6]$ } & Test acc. \\
    \midrule
    \midrule
ResNet-56 \cite{he2016deep}    & 0.85& 125 & 93.0\%  \\
CP-ResNet-56 \cite{he2017channel} & -  & 63 & 92.0\% \\
$\ell_1$ -ResNet-56 \cite{PrunningCornel2017} & 0.73 & 91 & 92.5\% \\
AMC-ResNet-56 \cite{he2018amc} & -  &63 &  91.9\% \\
Ghost-ResNet-56 
\cite{han2020ghostnet} & 0.43 & 63 & 92.7\% \\
\textbf{MGIC-ResNet-56 (ours)}  & 0.41 & 60 & \textbf{94.2\%}  \\
    \bottomrule
  \end{tabular}
  
\end{table}

\subsubsection{ImageNet}
The  ImageNet \cite{ImageNet} challenge ILSVRC 2012 consists of over $1.28$M training images and $50$K validation images from 1000 categories. We resize the images to $224\times 224$ \cite{he2016deep}.  We perform two experiments.

\begin{table}
  \caption{MGIC-ResNet50 architecture. MGIC-ResNet $(X)$ is the MGIC block applied with the ResNet bottleneck three-convolution block in Eq. \eqref{eq:bottleneck}, corresponding to the convolution kernel sizes in X as presented in \cite{he2016deep} (i.e., left numbers correspond to either $1\times1$ or $3\times 3$ filters of the convolutions, and the right numbers correspond to the output channels of each convolution). Conv2D is a 2D convolution layer followed by a BatchNorm operation and a ReLU non-linear activation. \# Rep is the number of block repetitions. $c_{out}$ denotes the number of output channels. A maxpool operation occurs after every convolution and MGIC layer.}
\label{tab:mgicresnet}
  \centering
 \begin{tabular}{lcccc}
    \toprule
        Input & Layer & \# Rep & $c_{out}$ \\

    \midrule
    \midrule
    $224^2 \times 3$ & Conv2D $7 \times 7$ & 1 & 64  \\
    $112^2 \times 64$ & Conv2D $3 \times 3$ & 1  &64  \\
    
    $112^2 \times 64$ & MGIC-ResNet $\left(\begin{bmatrix} 
1\times 1 \ , 64 \\
3\times 3 \ , 64 \\
1\times 1 \ , 256
\end{bmatrix} \right)$ & 3 & 256 \\ 

    $56^2 \times 256$ & MGIC-ResNet $\left(\begin{bmatrix} 
1\times 1 \ , 128\\
3\times 3 \ , 128 \\
1\times 1 \ , 512
\end{bmatrix} \right)$ & 4 & 512 \\ 

    $28^2 \times 512$ & MGIC-ResNet $\left(\begin{bmatrix} 
1\times 1 \ , 256\\
3\times 3 \ , 256 \\
1\times 1 \ , 1024
\end{bmatrix} \right)$ & 6 & 1024 \\ 
    $14^2 \times 1024$ & MGIC-ResNet $\left(\begin{bmatrix} 
1\times 1 \ , 512\\
3\times 3 \ , 512 \\
1\times 1 \ , 2048
\end{bmatrix} \right)$ & 3 & 2048 \\ 

$7^2 \times 2048$ & AvgPool2D $7 \times 7$ & 1 & 2048 \\
$1^2 \times 2048$ & FC & 1 & 1000 \\
    \bottomrule
  \end{tabular}
\end{table}

\subsubsection*{ResNet-50 compression} We compress the ResNet-50 architecture,  
and compare our MGIC approach with other methods. As the goal of this experiment is to compress a standard ResNet-50 \cite{he2016deep}, we follow the exact architecture of the latter, which includes input feature maps of widths of 256, 512, 1024 and 2048 channels. We only replace each ResNet block layer by an MGIC-ResNet block, denoted as MGIC$(\cdot)$, as depicted in Tab. \ref{tab:mgicresnet}. The results are reported in Table \ref{tab:ClassificationImageNet}, where we propose three variants of our MGIC-ResNet-50 network, differing in the $s_g$ parameter. For $s_c$ we chose 64, which lead to three to five levels in our MGIC blocks throughout the network. Our results outperform the rest of the considered methods, and our network with $s_g=64 , \ s_c=64$ also
outperforms ResNeXt-50 \cite{xie2017aggregated} ($25.0$M parameters, $4.2$B FLOPs, $77.8\%$ top-1 accuracy), which is not shown in the table because the ResNeXt architecture utilizes more channels than ResNet-50 and therefore is not directly comparable.

\begin{table}
\centering
  \caption{Comparison of state-of-the-art methods for compressing ResNet-50 on ImageNet dataset.}
  \label{tab:ClassificationImageNet}
  \begin{tabular}{lcccc}
    \toprule
    Model  & \makecell{Params\\ $[\times 10^6]$}& \makecell{FLOPs \\ $[\times 10^9]$ }  & \makecell{Top-1  \\ Acc.\%}  & \makecell{Top-5  \\ Acc.\%} \\
    \midrule
    \midrule 
    ResNet-50 \cite{he2016deep}  & 25.6 & 4.1 & 75.3  &  92.2 \\
    
    \midrule
     Thinet-ResNet-50 \cite{luo2017thinet} & 16.9  &2.6 & 72.1 & 90.3  \\
    
    NISP-ResNet-50-B \cite{yu2018nisp} & 14.4 & 2.3 & - & 90.8 \\
    
    Versatile-ResNet-50 \cite{wang2018learning} & 11.0 & 3.0 & 74.5  & 91.8 \\
    
    SSS-ResNet-50 \cite{huang2018data} & - & 2.8 &  74.2  & 91.9 \\

    Ghost-ResNet-50 \cite{han2020ghostnet} & 13.0 & 2.2 & 75.0  & 92.3 \\
    
    MGIC-ResNet-50 ($s_g=32, \ s_c=64$) (ours) & 9.4 & 1.6 & 75.8 & 92.9  \\
    
     MGIC-ResNet-50 ($s_g=64, \ s_c=64$) (ours) & 15.1 & 2.5 & \textbf{77.9} &\textbf{93.7}  \\
    
    \midrule
   Shift-ResNet-50 \cite{wu2018shift} &  6.0 & -  & 70.6  & 90.1 \\
   Taylor-FO-BN-ResNet-50 \cite{molchanov2019importance} &  7.9  & 1.3 &  71.7 &  - \\
   Slimmable-ResNet-50 0.5× \cite{yu2018slimmable} & 6.9 & 1.1 & 72.1 &  - \\
   MetaPruning-ResNet-50 \cite{liu2019metapruning} &  -  & 1.0 & 73.4  & - \\
   Ghost-ResNet-50 (s=4) \cite{han2020ghostnet} & 6.5 & 1.2 & 74.1 &  91.9 \\
   MGIC-ResNet-50 ($s_g=16,\ s_c=64$) (ours) & 6.2 & 1.0 & \textbf{74.3} & \textbf{92.0}  \\
    \bottomrule
  \end{tabular} 
\end{table}

\subsubsection*{ImageNet classification on a budget of FLOPs} 
\begin{figure}
    \centering
    \includegraphics[width=0.75\textwidth]{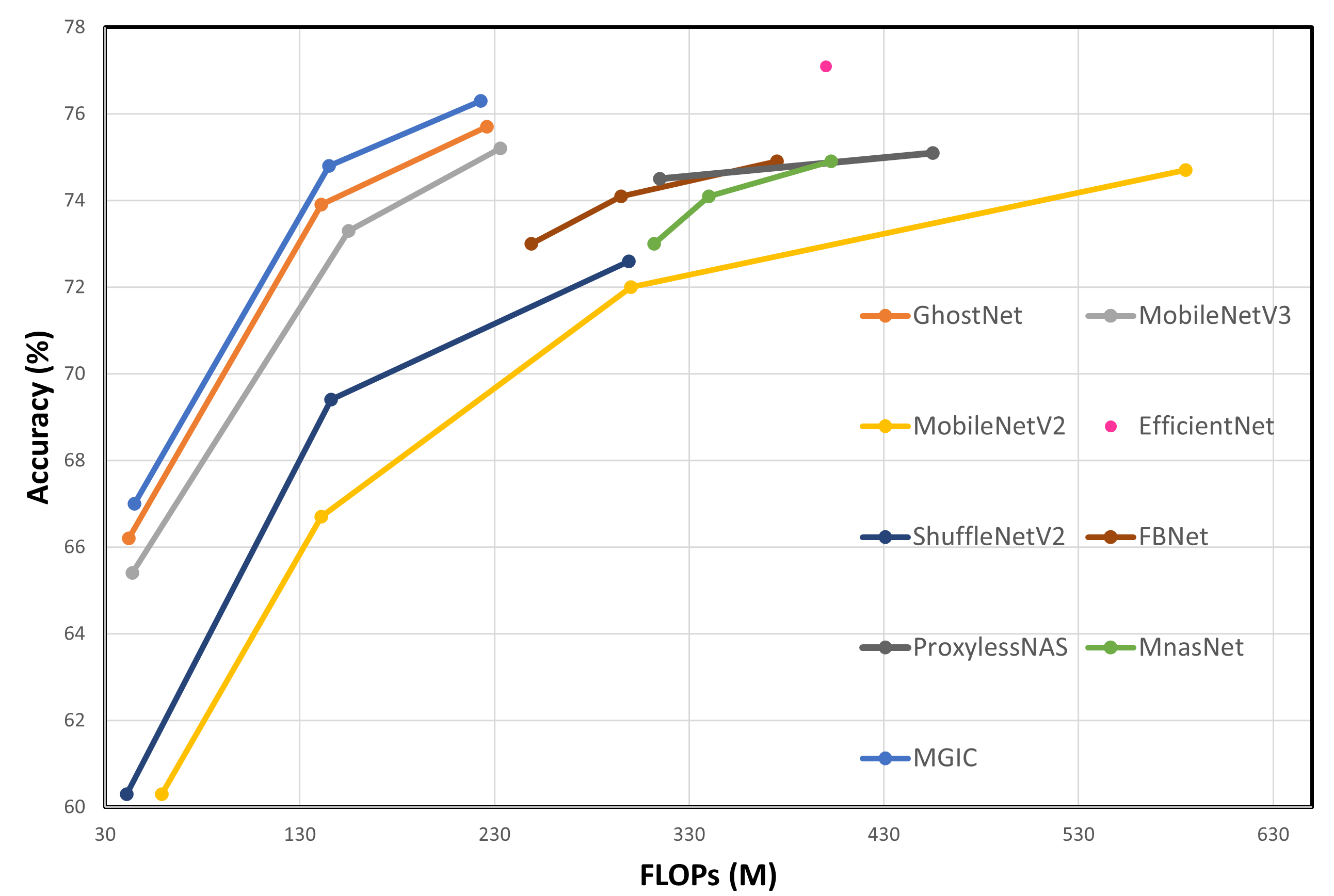}
    \caption{Top-1 accuracy v.s. FLOPs on ImageNet dataset.
    }
    \label{fig:accGraph}
\end{figure}

In this experiment we compare our approach with recent light networks. In particular, we follow the MobileNetV3-Large \cite{howard2019searching} architecture for its efficiency and high accuracy, and replace the standard MobileNetV3 block with our MGIC block.
 Our building blocks are MGIC-Bottlenecks (dubbed MGIC-bneck). That is, we build a MGIC version of the bottleneck from MobileNetV3. Our MGIC-MobileNetV3 is given in Tab. \ref{tab:mgicmobilenetv3}. Note, this is the $\times 1.0$ version, and can be modified via the width multiplier $\alpha$. Our parameter $s_g$ controls the group size -- therefore it determines the number of groups in each MGIC bottleneck. We denote the number of output channels by $c_{out}$ and the number of hidden channels within a block (the dimension of the square operator $K_{l_2}$ in a the block \eqref{eq:bottleneck}), also referred to as the expansion size, by $\# exp$. In case $c_{out}$ and $\# exp$ are not divisible by $s_g$, we set $s_g$ to the closest (smaller) integer to its intended value such that it divides them. For example, in our experiments we set $s_g=64$, and for the network defined in Tab. \ref{tab:mgicmobilenetv3}, the 5th MGIC-bneck layer has $\# exp = 120$ and $c_{out}=40$, meaning they do not divide by $64$. Therefore we modify $s_g$ to be the largest integer that is smaller than 64 and divides both $\# exp$ and $c_{out}$, giving $s_g=40$ in this example.
Our experiment is divided into three scales - small, medium, and large, where we scale our networks with width factors of $0.6, 1.0$ and $1.2$, respectively.
We find that our method obtains higher accuracy, with a similar number of FLOPs, as depicted from the results in Table \ref{tab:mobilenets} and Fig. \ref{fig:accGraph}. Specifically, we compare our methods with and without the use of the h--swish activation function \cite{howard2019searching}, where we see similar results. Compared to other popular and recent methods like MobileNetV3, GhostNet and ShuffleNetV2, we obtain better accuracy given the same FLOPs.

\begin{table}
  \caption{MGIC-MobileNetV$3$ architecture. MGIC-bneck denotes a MGIC-Bottleneck . The bottleneck is the same as in MobileNetV3, only in a multigrid-in-channels form. Conv2D is a 2D convolution layer followed by a BatchNorm operation and a ReLU non-linear activation. \# exp denotes the expansion size. $c_{out}$ denotes the number of output channels. SE stands for Squeeze-Excite. Pool denotes a maxpool operation, reducing the spatial size of the input .  - denotes a non-applicable option. \checkmark and $\times$ denote True and False, respectively.}
  \label{tab:mgicmobilenetv3}
  \centering
 \begin{tabular}{lcccccc}
    \toprule
    Input & Operation & \# exp &   $c_{out}$ & SE & Pool \\
    \midrule
    \midrule
    $224^2 \times 3$ &  Conv2D $3\times 3$ & 16 & - & $\times$ & \checkmark \\
    
    $112^2 \times 16$ &  MGIC-bneck & 16 & 16 & $\times$ & $\times$\\
    
    $112^2 \times 16$ &  MGIC-bneck & 48 & 24 & $\times$ & \checkmark \\
    
    $56^2 \times 24$ &  MGIC-bneck  & 72 & 24 & $\times$ & $\times$ \\
    
  $56^2 \times 24$ &  MGIC-bneck  & $72$ & $40$ & \checkmark & \checkmark \\
  
  $28^2 \times 40$ &  MGIC-bneck & $120$ & $40$ & \checkmark & $\times$ \\
  
    $28^2 \times 40$ &  MGIC-bneck & 240 & 80 & $\times$ & \checkmark \\
    
    $14^2 \times 80$ &  MGIC-bneck & 200 & 80 & $\times$ & $\times$ \\    

    $14^2 \times 80$ &  MGIC-bneck  & 184 & 80 & $\times$ & $\times$ \\    
    
    $14^2 \times 80$ &  MGIC-bneck  & 184 & 80 & $\times$ & $\times$ \\   

    $14^2 \times 80$ &  MGIC-bneck & 480 & 112 & \checkmark & $\times$ \\    

    $14^2 \times 112$ &  MGIC-bneck & 672 & 112 & \checkmark & $\times$ \\    

    $14^2 \times 112$ &  MGIC-bneck & 672 & 160 & \checkmark & \checkmark \\ 
    
    $7^2 \times 160$ &  MGIC-bneck & 960 & 160 & $\times$ & $\times$ \\    

    $7^2 \times 160$ & MGIC-bneck & 960 & 160 & \checkmark & $\times$ \\    
    
    $7^2 \times 160$ &  MGIC-bneck & 960 & 160 & $\times$ & $\times$  \\    
    
    $7^2 \times 160$ &  MGIC-bneck & 960 & 160 & \checkmark & $\times$ \\    
    
    \midrule
    
    $7^2 \times 160$ &  Conv2D $1\times 1$ & - & 960 & $\times$ & $\times$ \\ 
    
    $7^2 \times 960$ &  AvgPool $7\times 7$ & - & 960 & $\times$ & $\times$ \\ 
    
    $1^2 \times 960$ &  Conv2D $1\times 1$ & - & 1280 & $\times$ & $\times$ \\ 
    
    $1^2 \times 1280$ &  FC & - & 1000 & $\times$ & $\times$ \\ 
    \bottomrule
  \end{tabular}
\end{table}

\begin{table}
  \caption{Comparison of state-of-the-art light-weight networks on ImageNet dataset classification.}
  \label{tab:mobilenets}
  \centering
  \begin{tabular}{lcccc}
    \toprule
    Model  &\makecell{Params \\ $[\times 10^6]$} & \makecell{FLOPs \\ $[\times 10^6]$} & \makecell{Top-1 \\ Acc.\%}  &  \makecell{Top-5 \\ Acc.\%} \\
    \midrule
    \midrule 
ShuffleNetV1 0.5× (g=8) \cite{zhang2018shufflenet} & 1.0 & 40 & 58.8 &  81.0 \\
MobileNetV2 0.35× \cite{sandler2018mobilenetv2} & 1.7 & 59 & 60.3  &82.9 \\
ShuffleNetV2 0.5× \cite{ma2018shufflenet} & 1.4 & 41 & 61.1 & 82.6  \\
MobileNeXt 0.35× \cite{zhou2020rethinking} & 1.8 & 80 & 64.7 & - \\
MobileNetV3-Small 0.75× \cite{howard2019searching} & 2.4 & 44 & 65.4 & - \\
GhostNet 0.5× \cite{han2020ghostnet} & 2.6 & 42 & 66.2 &  86.6 \\
MGIC-MobileNetV3 0.6× (ours) & 2.3 & 48 & \textbf{67.0} & \textbf{87.3}   \\
MGIC-MobileNetV3 0.6× (ours) no h--swish & 2.3 & 45 & \textbf{66.8} & \textbf{86.9}   \\
\midrule
MobileNetV1 0.5× \cite{howard2017mobilenets} & 1.3 & 150 & 63.3 & 84.9 \\
MobileNetV2 0.6× \cite{sandler2018mobilenetv2} &  2.2 & 141 & 66.7 &  - \\
ShuffleNetV1 1.0× (g=3) \cite{zhang2018shufflenet} & 1.9 & 138 & 67.8 & 87.7 \\
ShuffleNetV2 1.0× \cite{ma2018shufflenet} & 2.3 & 146 & 69.4 &  88.9 \\
MobileNeXt 0.75× \cite{zhou2020rethinking} & 2.5 & 210 & 72.0 & - \\
MobileNetV3-Large 0.75× \cite{howard2019searching} & 4.0 & 155 & 73.3 & - \\
GhostNet 1.0× \cite{han2020ghostnet}  & 5.2 & 141 & 73.9 & 91.4 \\
MGIC-MobileNetV3 1.0×  (ours) & 5.2 & 145 & \textbf{74.8} & \textbf{92.0}  \\
MGIC-MobileNetV3 1.0×  (ours) no h--swish & 5.2 & 138 & \textbf{74.3} & \textbf{91.6}  \\
\midrule
MobileNetV2 1.0× \cite{sandler2018mobilenetv2} & 3.5 & 300 & 71.8 & 91.0 \\
ShuffleNetV2 1.5× \cite{ma2018shufflenet} & 3.5 & 299 & 72.6 & 90.6 \\
FE-Net 1.0× \cite{chen2019all} & 3.7 & 301 & 72.9 & - \\
FBNet-B \cite{wu2019fbnet} & 4.5 & 295 & 74.1 & - \\
ProxylessNAS \cite{cai2018proxylessnas} &  4.1  & 320 & 74.6 & 92.2 \\
MnasNet-A1 \cite{tan2019mnasnet} & 3.9 & 312 & 75.2 & 92.5 \\
MobileNeXt 1.0× \cite{zhou2020rethinking} & 3.4 & 300 & 74.0 & - \\
MobileNetV3-Large \cite{howard2019searching} 1.0× & 5.4 & 219 &  75.2 &  - \\
GhostNet 1.3× \cite{han2020ghostnet} & 7.3 & 226 & 75.7 & 92.7 \\ 
MGIC-MobileNetV3 1.2×  (ours) & 7.1 & 233 & \textbf{76.1} & \textbf{93.2}  \\
MGIC-MobileNetV3 1.2×  (ours) no h--swish & 7.1 & 217 & \textbf{76.2} & \textbf{93.4}  \\
    \bottomrule
  \end{tabular}
\end{table}

\vspace{5pt}
\paragraph{\textbf{Inference and training times}} We measure the single thread inference times on one $224 \times 224$ image using lightweight models on a Samsung Galaxy S8 mobile device (using the TFLite tool \cite{abadi2016tensorflow}), and an Intel i9-9820X CPU---see Tab. \ref{tab:runtimes} (averaged over 50 inferences). We observe that at least by these timings, the runtime of MGIC is on par with the considered architectures while obtaining higher accuracy. Additionally, we report the training time on a mini-batch of 128, $224 \times 224$ sized images from ImageNet, on an Nvidia Titan RTX GPU. Our MGIC requires slightly training time due to its higher complexity and multi-level in channels scheme. However, it yields higher accuracy compared to the considered methods.

\begin{table}[h!]
  \caption{Inference runtime of state-of-the-art small networks on a Samsung Galaxy S8 mobile device and a PC CPU and training time on an Nvidia Titan RTX GPU.}
  \label{tab:runtimes}
  \centering
  \begin{tabular}{lcccc}
    \toprule
    Metric  &\makecell{MobileNetV2 \\ 1.0x} & \makecell{MobileNetV3  \\ 0.75x}  & \makecell{GhostNet \\ 1.0x}  &  \makecell{MGIC \\ MobileNetV3 \\ 1.0x} \\
    \midrule
    \midrule 
Accuracy [\%]  & 71.8 & 73.3 & 73.9 & \textbf{74.8} \\
Mobile inference [ms]  & 795  & \textbf{418} &  487 & 480 \\
CPU inference [ms]  & 310  & \textbf{140} &  170 & 172 \\
\midrule
GPU training [s] & 0.262 & \textbf{0.215} & 0.237 & 0.311 \\

\bottomrule
\end{tabular}
\end{table}

\subsection{Image semantic segmentation}
We compare our method with MobileNetV3 on semantic segmentation on the Cityscapes \cite{cordts2016cityscapes} dataset. For the encoder part of the network, we build large and small variants, based on MobileNetV3-Large and MobileNetV3-Small, described in Tables 1-2 in \cite{howard2019searching}, respectively.
We also utilize the same LR-ASPP segmentation head and follow the observations from \cite{howard2019searching}. Namely, we reduce the number of channels in the last block of our networks by a factor of two and use $128$ filters in the segmentation head. For training, we use the same data augmentation and optimization approach as in \cite{chen2017rethinking}. The results are shown in Tab. \ref{tab:segmentation}, where report the mean intersection over union (mIoU) metric of our MGIC-Large with $s_g=64$ and $s_g=32$. We note that the results for the former are slightly better than those of MobileNetV3, while the performance of the latter are more favorble as they offer similar accuracy for less FLOPs and parameters. In addition, we read similar accuracy when using our MGIC-Small with$s_g=64$.

\begin{table}[ht!]
  \caption{Segmentation results on Cityscapes dataset. Metric is in mean intersection over union.} \label{tab:segmentation}
  \centering
 \begin{tabular}{lcccc}
    \toprule
    Backbone  & Params $[\times 10^6]$ & FLOPs $[\times 10^9]$  &  mIoU \% \\
    \midrule
    \midrule 
    MobileNetV$3$-Small & 0.47&  2.90 & 68.38 \\
    MGIC-Small $s_g=64$  (ours) & 0.48 & 2.73 & \textbf{68.52}  \\
    \hline
    MobileNetV$3$-Large &  1.51 & 9.74 & 72.64 \\
    MGIC-Large $s_g=32$ (ours)  &  1.32 & 8.87 & 71.02 \\
    MGIC-Large $s_g=64$  (ours)& 1.67 & 9.62 & \textbf{72.69}   \\
    \bottomrule
  \end{tabular}
\end{table}

%----------------------------------------------------------------------------------

\subsection{Point cloud classification}
The previous experiments were performed on structured CNNs, i.e., on 2D images. To further validate our method's generalization and usefulness, we incorporate it in graph convolutional networks (GCNs) to perform point cloud classification.
Specifically, we use a smaller version of the architecture from \cite{dgcnn}, where we alter the width of the last three classifier layers from $1024,512,256$ to $64$ in all of them. 
We define the architecture in Tab. \ref{tab:pointcloud}, where G-conv denotes a graph convolution layer, according to the methods listed in Tab. \ref{tab:gcn}, followed by a BatchNorm operation and a ReLU non-linear activation. MLP is realized by a simple $1 \times 1$ convolution followed by a BatchNorm operation and a ReLU non-linear activation. FC is a fully connected layer.
In all networks, we define the adjacency matrix using the k-NN algorithm with $k=10$ .Then, we replace the GCN block with each of the backbones listed in Tab. \ref{tab:gcn}, where we also report their performance on point-cloud classification on ModelNet-10 \cite{wu20153d} benchmark where we sample $1,024$ points from each shape.

\begin{table}[h!]
  \caption{Graph neural network for point cloud classification. G-conv is a graph convolution layer. MLP is a multi-layer perceptron. MaxPool is a global max-pooling layer. $c_{out}$ denotes the number of output channels. $\times$ denotes the number of repetitions of the respective layer.}
  \centering
  \label{tab:pointcloud}

 \begin{tabular}{lccc}
    \toprule
    Input & Operation &   $c_{out}$ \\
    \midrule
    \midrule
    $1024 \times 3 $ &  G-conv  & 64 \\
    $1024 \times 64 $ & 2 $ \times$ G-conv & 64 \\
    $1024 \times 64 $ &  MLP  & 64 \\
    $1024 \times 64 $ & 3 $ \times$ G-conv  & 64 \\
    $1024 \times 64 $ &  MLP  & 64 \\
    $1024 \times 64$ & MaxPool & 64 \\
    $1 \times 64 $ &  3 $ \times$ MLP  & 64 \\
    $1 \times 64 $ &  FC  & 10 \\
    \bottomrule
  \end{tabular}
\end{table}

\begin{table}[h!]
  \caption{ModelNet-10 classification.}
  \label{tab:gcn}
  \centering
 \begin{tabular}{lcccc}
    \toprule
    Backbone  & Params $[\times 10^6]$ & FLOPs $[\times 10^6]$  &  Accuracy \% \\
    \midrule
    \midrule 
    DGCNN \cite{dgcnn} &  0.16 & 125 & 91.6 \\
    diffGCN \cite{eliasof2020diffgcn} &  0.57 & 64 & 92.5 \\
   \textbf{MGIC-diffGCN (ours)} & \textbf{0.11}&  \textbf{13.7} & \textbf{92.9} \\
    \bottomrule
  \end{tabular}
 
\end{table}

%----------------------------------------------------------------------------------

\subsection{Ablation study} 

\paragraph{Parameter study} To determine the impact of the hyper-parameters $s_g$ and $s_c$, we experiment on CIFAR-10 and ImageNet data-sets for image classification. On CIFAR-10, we first fix $s_g$ to 16 and observe how the number of parameters, FLOPs, and accuracy of MGIC-ResNet-$56$ change. Secondly, we fix $s_c$ to 16, while modifying $s_g$, and examine our model's behavior, as reported in Tab. \ref{tab:FCCinfluence}-\ref{tab:imagenet_paramstudy}. On ImageNet, we examine two types of configurations using our MGIC-MobileNetV3 1.0x. In the first, we get $s_c = 64$ and experiment with different values of $s_g$. This experiment reveals the significance of the group size. Namely, it shows that as the group size grows, better accuracy is obtained (since it induces an increased channels connectivity), at the cost of more parameters. The second type of experiments considers various values of $s_c = s_g$, from 2 to 64. Like in the former, it can be concluded that increasing the coarsest grid size and the channel connectivity yields higher accuracy.  
Our conclusion from the results reported in Tab. \ref{tab:FCCinfluence}-\ref{tab:imagenet_paramstudy}, is that a growth of $s_g$ or $s_c$ yields better accuracy at the cost of more parameters and FLOPs, since an increased communication between the channels is allowed. However, by reducing $s_g$ and $s_c$, we obtain almost optimal accuracy at dramatically reduced costs, also as depicted in the experiments in Sec. \ref{sec:representation_experiment} -- \ref{sec:implicit_reconstruction_exp}. 

\begin{table}
  \caption{Influence of $s_c$ and $s_g$ in our MGIC framework on ResNet-56 architecture and CIFAR-10 dataset. $s_g$ is fixed to 16.}
  \label{tab:FCCinfluence}
  \centering
  \begin{tabular}{lcccc}
    \toprule
    $s_c$  & $s_g$ & Params $[\times 10^6]$ & FLOPs $[\times 10^6]$  &  Accuracy \% \\
    \midrule
    \midrule 
    64 & 16 & 0.53 & 91 & 94.7 \\
    32 & 16 & 0.5 & 76 & 94.6 \\
    16 & 16 & 0.47 & 65 & 94.3  \\
    \midrule
    \midrule 
    16 & 32 & 0.79 & 100 & 94.8 \\
    16 & 16 & 0.53 & 85 & 94.7  \\
    16 & 8 & 0.41 &  60 & 94.2 \\
    16 & 4 & 0.29 & 45 & 92.8  \\
    \bottomrule
  \end{tabular}
\end{table}

\begin{table}
  \caption{Hyper-parameter study on ImageNet}
  \label{tab:imagenet_paramstudy}
  \centering
  \begin{tabular}{lccc}
    \toprule
    $s_c$  & $s_g$ & Params $[\times 10^6]$  &  Accuracy \% \\
    \midrule
    \midrule
     64 & 2  & 4.2 & 70.2 \\
     64 & 4 & 4.2 & 70.6 \\
     64 & 8 & 4.3 & 71.6 \\
     64 & 16 & 4.5 & 71.9 \\
     64 & 32 & 4.8 & 73.2 \\
     64 & 64 & 5.2 & 74.8 \\
     \midrule
     2 & 2 & 4.2 & 69.9 \\ 
     4 & 4 & 4.2 & 70.5 \\
     8 & 8 & 4.3 & 70.9 \\
     16 & 16 & 4.5 & 71.6 \\
     32 & 32 & 4.9 & 72.7 \\
    \bottomrule
  \end{tabular}
\end{table}

\paragraph{Training of MGIC}

\begin{figure}
    \centering
    \includegraphics[width=0.6\textwidth]{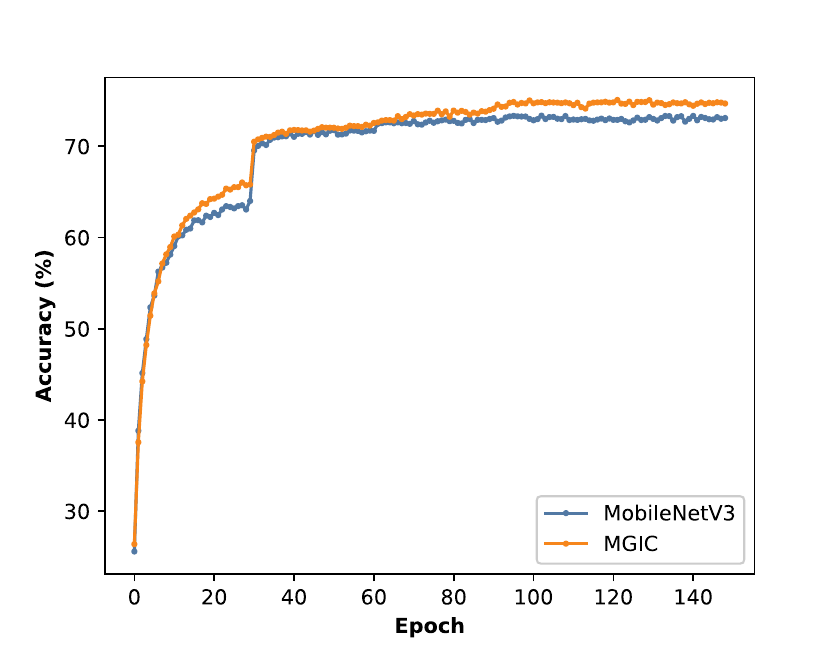}
    \caption{Per-epoch accuracy ($\%$) on ImageNet classification. With MobileNetV3 the best accuracy  is 73.34\% and 73.36\% throughout the first 100 and 150 epochs, respectively. With MGIC-MobileNetV3, the corresponding accuracy reads are 75.02\% and 75.08\%, respectively.}  
    \label{fig:convergence}
\end{figure}

Another interesting aspect is whether the performance improvement of MGIC over MobileNetV3 stems from higher efficacy of MGIC (per parameter or FLOP), or an easier and better training. In the case of the latter, it is expected that training MobileNetV3 for more epochs will lead to a smaller performance gap compared to our MGIC. In Fig. \ref{fig:convergence}, we present the obtained accuracy per-epoch on the ImageNet data-set with 150 epochs (as opposed to the 100 epochs results that are reported in Tab. \ref{tab:ClassificationImageNet}.)
It can be seen from Fig. \ref{fig:convergence} that adding more epochs, using the same training policy, does not improve performance for any of the methods. Therefore, it seems that given our training policy (which is identical to the one used in MobileNetV3), our MGIC shows better efficacy rather than easier or better training.

%--------------------------------------------------------------------

\section{Conclusion}
We present a novel multigrid-in-channels (MGIC) approach that improves the efficiency of convolutional neural networks (CNN) both in parameters and FLOPs, while using easy-to-implement structured grouped convolutions in the channel space. Applying MGIC, we achieve full coupling through a multilevel hierarchy of the channels, at only $\CO(c)$ cost, unlike standard convolution layers that require $\CO(c^2)$. This property is significant and desired both to reduce training and inference times, which also translates to a reduction in energy consumption. We also note that MGIC is most beneficial for wide networks, which are usually favored for state-of-the-art accuracy and performance.
Our experiments for various tasks suggest that MGIC achieves comparable or superior accuracy than other recent light-weight architectures at a given budget.
Our MGIC block offers a universal approach for producing lightweight versions of networks suitable for different kinds of CNNs, GCNs, and traditional NNs, where fully-connected layers are applied. Furthermore, it is future-ready, meaning it can also compress future architectures when available.

\section*{Acknowledgments}
The research reported in this paper was supported by the Israel Innovation Authority through Avatar
consortium, and by grant no. 2018209 from the United States - Israel
Binational Science Foundation (BSF), Jerusalem, Israel.  ME
is supported by Kreitman High-tech scholarship.
\bibliography{deepCNNs}

\begin{thebibliography}{10}

\bibitem{abadi2016tensorflow}
{\sc M.~Abadi, P.~Barham, J.~Chen, Z.~Chen, A.~Davis, J.~Dean, M.~Devin,
  S.~Ghemawat, G.~Irving, M.~Isard, et~al.}, {\em Tensorflow: A system for
  large-scale machine learning}, in 12th $\{$USENIX$\}$ symposium on operating
  systems design and implementation ($\{$OSDI$\}$ 16), 2016, pp.~265--283.

\bibitem{bar2019unsupervised}
{\sc L.~Bar and N.~Sochen}, {\em Strong solutions for pde-based tomography by
  unsupervised learning}, SIAM Journal on Imaging Sciences, 14 (2021),
  pp.~128--155.

\bibitem{bianchini2014complexity}
{\sc M.~Bianchini and F.~Scarselli}, {\em On the complexity of neural network
  classifiers: A comparison between shallow and deep architectures}, IEEE
  transactions on neural networks and learning systems, 25 (2014),
  pp.~1553--1565.

\bibitem{borzi2009multigrid}
{\sc A.~Borzi and V.~Schulz}, {\em Multigrid methods for pde optimization},
  SIAM review, 51 (2009), pp.~361--395.

\bibitem{brandt2003multigrid}
{\sc A.~Brandt and D.~Ron}, {\em Multigrid solvers and multilevel optimization
  strategies}, in Multilevel optimization in VLSICAD, Springer, 2003,
  pp.~1--69.

\bibitem{briggs2000multigrid}
{\sc W.~L. Briggs, V.~E. Henson, and S.~F. McCormick}, {\em A multigrid
  tutorial}, SIAM, 2000.

\bibitem{brown2020languagegpt}
{\sc T.~Brown, B.~Mann, N.~Ryder, M.~Subbiah, J.~D. Kaplan, P.~Dhariwal,
  A.~Neelakantan, P.~Shyam, G.~Sastry, A.~Askell, et~al.}, {\em Language models
  are few-shot learners}, Advances in neural information processing systems, 33
  (2020), pp.~1877--1901.

\bibitem{cai2018proxylessnas}
{\sc H.~Cai, L.~Zhu, and S.~Han}, {\em Proxyless{NAS}: Direct neural
  architecture search on target task and hardware}, in International Conference
  on Learning Representations, 2019,
  \url{https://arxiv.org/pdf/1812.00332.pdf}.

\bibitem{chang2017multi}
{\sc B.~Chang, L.~Meng, E.~Haber, F.~Tung, and D.~Begert}, {\em Multi-level
  residual networks from dynamical systems view}, arXiv preprint
  arXiv:1710.10348,  (2017).

\bibitem{SparsConvCornell2017}
{\sc S.~Changpinyo, M.~Sandler, and A.~Zhmoginov}, {\em The power of sparsity
  in convolutional neural networks}, 2017,
  \url{https://arxiv.org/abs/1702.06257}.

\bibitem{chen2018big}
{\sc C.-F.~R. Chen, Q.~Fan, N.~Mallinar, T.~Sercu, and R.~Feris}, {\em
  {Big-Little Net: An Efficient Multi-Scale Feature Representation for Visual
  and Speech Recognition}}, in International Conference on Learning
  Representations, 2019, \url{https://openreview.net/forum?id=HJMHpjC9Ym}.

\bibitem{chen2017rethinking}
{\sc L.-C. Chen, G.~Papandreou, F.~Schroff, and H.~Adam}, {\em Rethinking
  atrous convolution for semantic image segmentation}, arXiv preprint
  arXiv:1706.05587,  (2017).

\bibitem{chen2019all}
{\sc W.~Chen, D.~Xie, Y.~Zhang, and S.~Pu}, {\em All you need is a few shifts:
  Designing efficient convolutional neural networks for image classification},
  in Proceedings of the IEEE Conference on Computer Vision and Pattern
  Recognition, 2019, pp.~7241--7250.

\bibitem{Chen:2019tp}
{\sc Y.~Chen, H.~Fan, B.~Xu, Z.~Yan, Y.~Kalantidis, M.~Rohrbach, S.~Yan, and
  J.~Feng}, {\em Drop an {Octave}: {Reducing} spatial redundancy in
  convolutional neural networks with octave convolution}, in Proceedings of the
  IEEE International Conference on Computer Vision (ICCV), 2019,
  pp.~3435--3444.

\bibitem{cordts2016cityscapes}
{\sc M.~Cordts, M.~Omran, S.~Ramos, T.~Rehfeld, M.~Enzweiler, R.~Benenson,
  U.~Franke, S.~Roth, and B.~Schiele}, {\em The cityscapes dataset for semantic
  urban scene understanding}, in Proc. of the IEEE Conference on Computer
  Vision and Pattern Recognition (CVPR), 2016.

\bibitem{de2008multilevel}
{\sc H.~De~Sterck, T.~A. Manteuffel, S.~F. McCormick, Q.~Nguyen, and J.~Ruge},
  {\em Multilevel adaptive aggregation for markov chains, with application to
  web ranking}, SIAM Journal on Scientific Computing, 30 (2008),
  pp.~2235--2262.

\bibitem{ImageNet}
{\sc J.~Deng, W.~Dong, R.~Socher, L.-J. Li, K.~Li, and L.~Fei-Fei}, {\em
  {ImageNet: A Large-Scale Hierarchical Image Database}}, in CVPR09, 2009.

\bibitem{eliasof2020diffgcn}
{\sc M.~Eliasof and E.~Treister}, {\em {DiffGCN}: Graph convolutional networks
  via differential operators and algebraic multigrid pooling}, Advances in
  Neural Information Processing Systems (NeurIPS),  (2020).

\bibitem{ephrath2020leanconvnets}
{\sc J.~Ephrath, M.~Eliasof, L.~Ruthotto, E.~Haber, and E.~Treister}, {\em
  {LeanConvNets}: Low-cost yet effective convolutional neural networks}, IEEE
  Journal of Selected Topics in Signal Processing,  (2020).

\bibitem{falgout2014parallel}
{\sc R.~D. Falgout, S.~Friedhoff, T.~V. Kolev, S.~P. MacLachlan, and J.~B.
  Schroder}, {\em Parallel time integration with multigrid}, SIAM Journal on
  Scientific Computing, 36 (2014), pp.~C635--C661.

\bibitem{girshick2014rich}
{\sc R.~Girshick, J.~Donahue, T.~Darrell, and J.~Malik}, {\em Rich feature
  hierarchies for accurate object detection and semantic segmentation}, in
  Proceedings of the IEEE conference on computer vision and pattern
  recognition, 2014, pp.~580--587.

\bibitem{guenther2020layer}
{\sc S.~G{\"u}nther, L.~Ruthotto, J.~B. Schroder, E.~C. Cyr, and N.~R. Gauger},
  {\em Layer-parallel training of deep residual neural networks}, SIAM Journal
  on Mathematics of Data Science, 2 (2020), pp.~1--23.

\bibitem{haber2018learning}
{\sc E.~Haber, L.~Ruthotto, E.~Holtham, and S.-H. Jun}, {\em Learning across
  scales---multiscale methods for convolution neural networks}, in
  Thirty-Second AAAI Conference on Artificial Intelligence, 2018.

\bibitem{hammernik2018learning}
{\sc K.~Hammernik, T.~Klatzer, E.~Kobler, M.~P. Recht, D.~K. Sodickson,
  T.~Pock, and F.~Knoll}, {\em Learning a variational network for
  reconstruction of accelerated mri data}, Magnetic resonance in medicine, 79
  (2018), pp.~3055--3071.

\bibitem{han2020ghostnet}
{\sc K.~Han, Y.~Wang, Q.~Tian, J.~Guo, C.~Xu, and C.~Xu}, {\em {GhostNet}: More
  features from cheap operations}, in Proceedings of the IEEE/CVF Conference on
  Computer Vision and Pattern Recognition, 2020, pp.~1580--1589.

\bibitem{SparseReguSongHan2016}
{\sc S.~Han, J.~Pool, S.~Narang, H.~Mao, E.~Gong, S.~Tang, E.~Elsen, P.~Vajda,
  M.~Paluri, J.~Tran, B.~Catanzaro, and W.~J. Dally}, {\em {DSD}:
  Dense-sparse-dense training for deep neural networks}, in Proceedings of the
  International Conference on Learning Representations (ICLR), 2017.

\bibitem{SongHan2015}
{\sc S.~Han, J.~Pool, J.~Tran, and W.~J. Dally}, {\em Learning both weights and
  connections for efficient neural network}, International Journal of Computer
  Vision, 5 (2015), pp.~1135--1143.

\bibitem{pruning92}
{\sc B.~Hassibi and D.~G. Stork}, {\em Second order derivatives for network
  pruning: Optimal brain surgeon reconstruction}, International Journal of
  Computer Vision, 5 (1992), pp.~164--171.

\bibitem{he2019mgnet}
{\sc J.~He and J.~Xu}, {\em Mgnet: A unified framework of multigrid and
  convolutional neural network}, Science China mathematics, 62 (2019),
  pp.~1331--1354.

\bibitem{he2016deep}
{\sc K.~He, X.~Zhang, S.~Ren, and J.~Sun}, {\em Deep residual learning for
  image recognition}, in Proceedings of the IEEE Conference on Computer Vision
  and Pattern Recognition, 2016, pp.~770--778.

\bibitem{he2018amc}
{\sc Y.~He, J.~Lin, Z.~Liu, H.~Wang, L.-J. Li, and S.~Han}, {\em Amc: Automl
  for model compression and acceleration on mobile devices}, in Proceedings of
  the European Conference on Computer Vision (ECCV), 2018, pp.~784--800.

\bibitem{he2017channel}
{\sc Y.~He, X.~Zhang, and J.~Sun}, {\em Channel pruning for accelerating very
  deep neural networks}, in Proceedings of the IEEE International Conference on
  Computer Vision, 2017, pp.~1389--1397.

\bibitem{HochreiterLearning}
{\sc S.~Hochreiter, A.~Younger, and P.~Conwell}, {\em Learning to learn using
  gradient descent}, 09 2001, pp.~87--94,
  \url{https://doi.org/10.1007/3-540-44668-0_13}.

\bibitem{howard2019searching}
{\sc A.~Howard, M.~Sandler, G.~Chu, L.-C. Chen, B.~Chen, M.~Tan, W.~Wang,
  Y.~Zhu, R.~Pang, V.~Vasudevan, et~al.}, {\em Searching for {MobileNetv3}}, in
  Proceedings of the IEEE International Conference on Computer Vision, 2019,
  pp.~1314--1324.

\bibitem{howard2017mobilenets}
{\sc A.~G. Howard, M.~Zhu, B.~Chen, D.~Kalenichenko, W.~Wang, T.~Weyand,
  M.~Andreetto, and H.~Adam}, {\em {MobileNets}: Efficient convolutional neural
  networks for mobile vision applications}, arXiv preprint arXiv:1704.04861,
  (2017).

\bibitem{huang2019gpipe}
{\sc Y.~Huang, Y.~Cheng, A.~Bapna, O.~Firat, D.~Chen, M.~Chen, H.~Lee,
  J.~Ngiam, Q.~V. Le, Y.~Wu, et~al.}, {\em {Gpipe}: Efficient training of giant
  neural networks using pipeline parallelism}, in Advances in Neural
  Information Processing Systems, 2019, pp.~103--112.

\bibitem{huang2018data}
{\sc Z.~Huang and N.~Wang}, {\em Data-driven sparse structure selection for
  deep neural networks}, in Proceedings of the European conference on computer
  vision (ECCV), 2018, pp.~304--320.

\bibitem{MultigridNN}
{\sc T.-W. Ke, M.~Maire, and S.~X. Yu}, {\em Multigrid neural architectures},
  in Proceedings of the IEEE Conference on Computer Vision and Pattern
  Recognition (CVPR), 2017, pp.~6665--6673.

\bibitem{khoo2018switchnet}
{\sc Y.~Khoo and L.~Ying}, {\em {SwitchNet}: a neural network model for forward
  and inverse scattering problems}, SIAM Journal on Scientific Computing, 41
  (2019), pp.~A3182--A3201.

\bibitem{kopanivcakova2021globally}
{\sc A.~Kopani{\v{c}}{\'a}kov{\'a} and R.~Krause}, {\em Globally convergent
  multilevel training of deep residual networks}, arXiv preprint
  arXiv:2107.07572,  (2021).

\bibitem{krizhevsky2009learning}
{\sc A.~Krizhevsky and G.~Hinton}, {\em Learning multiple layers of features
  from tiny images},  (2009).

\bibitem{krizhevsky2012imagenet}
{\sc A.~Krizhevsky, I.~Sutskever, and G.~E. Hinton}, {\em Imagenet
  classification with deep convolutional neural networks}, in Advances in
  Neural Information Processing Systems, 2012, pp.~1097--1105.

\bibitem{LeCun1990}
{\sc Y.~LeCun, B.~E. Boser, and J.~S. Denker}, {\em {Handwritten digit
  recognition with a back-propagation network}}, in Advances in neural
  information processing systems, 1990, pp.~396--404.

\bibitem{PrunningCornel2017}
{\sc H.~Li, A.~Kadav, I.~Durdanovic, H.~Samet, and H.~P. Graf}, {\em Pruning
  filters for efficient {ConvNets}}, in Proceedings of the International
  Conference on Learning Representations (ICLR), 2017.

\bibitem{liu2019metapruning}
{\sc Z.~Liu, H.~Mu, X.~Zhang, Z.~Guo, X.~Yang, K.-T. Cheng, and J.~Sun}, {\em
  Metapruning: Meta learning for automatic neural network channel pruning}, in
  Proceedings of the IEEE International Conference on Computer Vision, 2019,
  pp.~3296--3305.

\bibitem{livne2012lean}
{\sc O.~E. Livne and A.~Brandt}, {\em Lean algebraic multigrid (lamg): Fast
  graph laplacian linear solver}, SIAM Journal on Scientific Computing, 34
  (2012), pp.~B499--B522.

\bibitem{lu2020deepxde}
{\sc L.~Lu, X.~Meng, Z.~Mao, and G.~E. Karniadakis}, {\em Deepxde: A deep
  learning library for solving differential equations}, SIAM Review, 63 (2021),
  pp.~208--228.

\bibitem{luo2017thinet}
{\sc J.-H. Luo, J.~Wu, and W.~Lin}, {\em Thinet: A filter level pruning method
  for deep neural network compression}, in Proceedings of the IEEE
  International Conference on Computer Vision (ICCV), 2017, pp.~5058--5066.

\bibitem{ma2018shufflenet}
{\sc N.~Ma, X.~Zhang, H.-T. Zheng, and J.~Sun}, {\em {ShuffleNet V2}: Practical
  guidelines for efficient {CNN} architecture design}, in Proceedings of the
  European Conference on Computer Vision (ECCV), 2018, pp.~116--131.

\bibitem{molchanov2019importance}
{\sc P.~Molchanov, A.~Mallya, S.~Tyree, I.~Frosio, and J.~Kautz}, {\em
  Importance estimation for neural network pruning}, in Proceedings of the IEEE
  Conference on Computer Vision and Pattern Recognition, 2019,
  pp.~11264--11272.

\bibitem{molchanov2016pruning}
{\sc P.~Molchanov, S.~Tyree, T.~Karras, T.~Aila, and J.~Kautz}, {\em Pruning
  convolutional neural networks for resource efficient transfer learning}, in
  Proceedings of the International Conference on Learning Representations
  (ICLR), 2017.

\bibitem{napov2017efficient}
{\sc A.~Napov and Y.~Notay}, {\em An efficient multigrid method for graph
  laplacian systems ii: robust aggregation}, SIAM journal on scientific
  computing, 39 (2017), pp.~S379--S403.

\bibitem{pang2019fpinns}
{\sc G.~Pang, L.~Lu, and G.~E. Karniadakis}, {\em fpinns: Fractional
  physics-informed neural networks}, SIAM Journal on Scientific Computing, 41
  (2019), pp.~A2603--A2626.

\bibitem{park2019deepsdf}
{\sc J.~J. Park, P.~Florence, J.~Straub, R.~Newcombe, and S.~Lovegrove}, {\em
  Deepsdf: Learning continuous signed distance functions for shape
  representation}, in Proceedings of the IEEE/CVF Conference on Computer Vision
  and Pattern Recognition, 2019, pp.~165--174.

\bibitem{paszke2017automatic}
{\sc A.~Paszke, S.~Gross, S.~Chintala, G.~Chanan, E.~Yang, Z.~DeVito, Z.~Lin,
  A.~Desmaison, L.~Antiga, and A.~Lerer}, {\em Automatic differentiation in
  pytorch}, in Advances in Neural Information Processing Systems, 2017.

\bibitem{pelt2018mixed}
{\sc D.~M. Pelt and J.~A. Sethian}, {\em A mixed-scale dense convolutional
  neural network for image analysis}, Proceedings of the National Academy of
  Sciences, 115 (2018), pp.~254--259.

\bibitem{raissi2019physics}
{\sc M.~Raissi, P.~Perdikaris, and G.~E. Karniadakis}, {\em Physics-informed
  neural networks: A deep learning framework for solving forward and inverse
  problems involving nonlinear partial differential equations}, Journal of
  Computational Physics, 378 (2019), pp.~686--707.

\bibitem{sandler2018mobilenetv2}
{\sc M.~Sandler, A.~Howard, M.~Zhu, A.~Zhmoginov, and L.-C. Chen}, {\em
  {MobileNetV2}: Inverted residuals and linear bottlenecks}, in Proceedings of
  the IEEE Conference on Computer Vision and Pattern Recognition, 2018,
  pp.~4510--4520.

\bibitem{szegedy2015going}
{\sc C.~Szegedy, W.~Liu, Y.~Jia, P.~Sermanet, S.~Reed, D.~Anguelov, D.~Erhan,
  V.~Vanhoucke, and A.~Rabinovich}, {\em Going deeper with convolutions}, in
  Proceedings of the IEEE conference on computer vision and pattern
  recognition, 2015, pp.~1--9.

\bibitem{tan2019mnasnet}
{\sc M.~Tan, B.~Chen, R.~Pang, V.~Vasudevan, M.~Sandler, A.~Howard, and Q.~V.
  Le}, {\em Mnasnet: Platform-aware neural architecture search for mobile}, in
  Proceedings of the IEEE Conference on Computer Vision and Pattern
  Recognition, 2019, pp.~2820--2828.

\bibitem{tan2019efficientnet}
{\sc M.~Tan and Q.~V. Le}, {\em {EfficientNet}: Rethinking model scaling for
  convolutional neural networks}, in International Conference on Machine
  Learning (ICML), 2019.

\bibitem{treister2016multilevel}
{\sc E.~Treister, J.~S. Turek, and I.~Yavneh}, {\em A multilevel framework for
  sparse optimization with application to inverse covariance estimation and
  logistic regression}, SIAM Journal on Scientific Computing, 38 (2016),
  pp.~S566--S592.

\bibitem{vanvek1996algebraic}
{\sc P.~Van{\v{e}}k, J.~Mandel, and M.~Brezina}, {\em Algebraic multigrid by
  smoothed aggregation for second and fourth order elliptic problems},
  Computing, 56 (1996), pp.~179--196.

\bibitem{wang2019elastic}
{\sc H.~Wang, A.~Kembhavi, A.~Farhadi, A.~L. Yuille, and M.~Rastegari}, {\em
  Elastic: Improving cnns with dynamic scaling policies}, in Proceedings of the
  IEEE Conference on Computer Vision and Pattern Recognition (CVPR), 2019,
  pp.~2258--2267.

\bibitem{dgcnn}
{\sc Y.~Wang, Y.~Sun, Z.~Liu, S.~E. Sarma, M.~M. Bronstein, and J.~M. Solomon},
  {\em Dynamic graph cnn for learning on point clouds}, ACM Transactions on
  Graphics (TOG),  (2019).

\bibitem{wang2018learning}
{\sc Y.~Wang, C.~Xu, X.~Chunjing, C.~Xu, and D.~Tao}, {\em Learning versatile
  filters for efficient convolutional neural networks}, in Advances in Neural
  Information Processing Systems, 2018, pp.~1608--1618.

\bibitem{wu2019fbnet}
{\sc B.~Wu, X.~Dai, P.~Zhang, Y.~Wang, F.~Sun, Y.~Wu, Y.~Tian, P.~Vajda,
  Y.~Jia, and K.~Keutzer}, {\em Fbnet: Hardware-aware efficient convnet design
  via differentiable neural architecture search}, in Proceedings of the IEEE
  Conference on Computer Vision and Pattern Recognition, 2019,
  pp.~10734--10742.

\bibitem{wu2018shift}
{\sc B.~Wu, A.~Wan, X.~Yue, P.~Jin, S.~Zhao, N.~Golmant, A.~Gholaminejad,
  J.~Gonzalez, and K.~Keutzer}, {\em Shift: A zero flop, zero parameter
  alternative to spatial convolutions}, in Proceedings of the IEEE Conference
  on Computer Vision and Pattern Recognition, 2018, pp.~9127--9135.

\bibitem{wu20153d}
{\sc Z.~Wu, S.~Song, A.~Khosla, F.~Yu, L.~Zhang, X.~Tang, and J.~Xiao}, {\em 3d
  shapenets: A deep representation for volumetric shapes}, in Proceedings of
  the IEEE conference on computer vision and pattern recognition, 2015,
  pp.~1912--1920.

\bibitem{xie2017aggregated}
{\sc S.~Xie, R.~Girshick, P.~Doll{\'a}r, Z.~Tu, and K.~He}, {\em Aggregated
  residual transformations for deep neural networks}, in Proceedings of the
  IEEE conference on computer vision and pattern recognition, 2017,
  pp.~1492--1500.

\bibitem{yu2018slimmable}
{\sc J.~Yu, L.~Yang, N.~Xu, J.~Yang, and T.~Huang}, {\em Slimmable neural
  networks}, arXiv preprint arXiv:1812.08928,  (2018).

\bibitem{yu2018nisp}
{\sc R.~Yu, A.~Li, C.-F. Chen, J.-H. Lai, V.~I. Morariu, X.~Han, M.~Gao, C.-Y.
  Lin, and L.~S. Davis}, {\em Nisp: Pruning networks using neuron importance
  score propagation}, in Proceedings of the IEEE Conference on Computer Vision
  and Pattern Recognition, 2018, pp.~9194--9203.

\bibitem{Zagoruyko:2016wo}
{\sc S.~Zagoruyko and N.~Komodakis}, {\em Wide residual networks}, in
  Proceedings of the British Machine Vision Conference (BMVC), E.~R.~H. Richard
  C.~Wilson and W.~A.~P. Smith, eds., BMVA Press, September 2016,
  pp.~87.1--87.12, \url{https://doi.org/10.5244/C.30.87},
  \url{https://dx.doi.org/10.5244/C.30.87}.

\bibitem{zhang2018shufflenet}
{\sc X.~Zhang, X.~Zhou, M.~Lin, and J.~Sun}, {\em {ShuffleNet}: An extremely
  efficient convolutional neural network for mobile devices}, in Proceedings of
  the IEEE Conference on Computer Vision and Pattern Recognition, 2018,
  pp.~6848--6856.

\bibitem{zhou2020rethinking}
{\sc D.~Zhou, Q.~Hou, Y.~Chen, J.~Feng, and S.~Yan}, {\em Rethinking bottleneck
  structure for efficient mobile network design}, ECCV, August, 2 (2020).

\end{thebibliography}
\bibliographystyle{siamplain}
\end{document}